\documentclass[11pt]{article}
\usepackage[margin=1in]{geometry}
\usepackage{authblk}
\usepackage{amsmath,amssymb,amsthm,mathrsfs}
\usepackage{mathtools}
\usepackage{graphicx}
\usepackage[ round, sort]{natbib}
\usepackage{enumitem}
\usepackage{setspace}
\usepackage{subfigure}
\usepackage{bbm}
\usepackage{bm}
\usepackage{booktabs}
\usepackage{caption}
\usepackage{color}
\usepackage{algorithm}
\usepackage{algorithmic}
\usepackage{graphicx}
\usepackage{multirow}
\usepackage{diagbox}

\RequirePackage[colorlinks,citecolor=blue,urlcolor=blue]{hyperref}
\DeclareMathOperator*{\argmax}{argmax}
\newtheorem{theorem}{Theorem}[section]
\newtheorem{lemma}{Lemma}[section]
\newtheorem{proposition}{Proposition}[section]

\theoremstyle{definition}
\numberwithin{equation}{section}

\begin{document}
\parskip=8pt
\baselineskip=16.8pt
\title{Stochastically Constrained Best Arm Identification with Thompson Sampling}
\author[1]{Le Yang\thanks{lyang272-c@my.cityu.edu.hk}}
\author[1]{Siyang Gao\thanks{siyangao@cityu.edu.hk}}
\author[2]{Cheng Li\thanks{stalic@nus.edu.sg}}
\author[3]{Yi Wang\thanks{yiwang@eee.hku.hk}}
\affil[1]{Department of Systems Engineering, City University of Hong Kong}
\affil[2]{Department of Statistics and Data Science, National University of Singapore}
\affil[3]{Department of Electrical and Electronic Engineering, The University of Hong Kong}
\date{}
\maketitle	
		
\begin{abstract}                          
		We consider the problem of the best arm identification in the presence of stochastic constraints, where there is a finite number of arms associated with multiple performance measures. The goal is to identify the arm that optimizes the objective measure subject to constraints on the remaining measures. We will explore the popular idea of Thompson sampling (TS) as a means to solve it. To the best of our knowledge, it is the first attempt to extend TS to this problem. We will design a TS-based sampling algorithm, establish its asymptotic optimality in the rate of posterior convergence, and demonstrate its superior performance using numerical examples.
\end{abstract}
		
\section{Introduction}

In the best arm identification (BAI) problem, there is a finite number of arms with unknown mean performances. An agent sequentially chooses an arm to pull and observes a noisy reward sample of it. At the end of the sampling stage, the agent selects the arm which he/she believes to be the best, i.e., the one with the largest mean reward. The BAI has a long history dating back to the 50s \citep*{bechhofer1954tow,bechhofer1958sequential} and has been widely studied in the machine learning \citep*{even2006action,yang2024improving} and simulation communities \citep*{chen1997new,lee2010review,gao2016optimal,li2023convergence}.

However, in many real applications, the reward from pulling an arm is a random vector instead of a scalar. It is of interest to optimize an (important) objective measure in the reward vector while putting the rest measures in the constraints, i.e., the agent wants to identify the arm with the largest objective mean reward among alternatives whose non-objective measures satisfy given constraints. For example, in clinical trials, we might want to select the most effective drug from a few alternatives while keeping the probability of the drug causing adverse reactions within a reasonable range. In financial lending, we might be interested in designing a classifier that maximizes the prediction accuracy, and in the meantime the probability of unfairly predicting a high credit risk for some subgroups of the population applying for a loan should not exceed a certain threshold. In these applications, the agent aims to identify the best feasible arm among a finite set of alternatives. We call it the best feasible arm identification (BFAI).

In this paper, we focus on the fixed-budget setting of BFAI, in which the total number of samples (budget) is fixed and known by the agent. The goal is to minimize the probability of selecting an arm that is not best feasible, i.e., the probability of false selection (PFS).

To solve it, we adopt the basic idea of Thompson Sampling (TS), which has proven highly effective for bandit problems. An intuitive extension of it to BFAI would be to pull the arm according to the posterior probability of it being the best feasible. Although this idea is quite straightforward, through some in-depth analysis, we found that it tends to allocate too many samples to the best feasible arm in the long run, which could seriously compromise the algorithm's convergence and empirical performance. To fix it, we introduce a parameter $\beta$ to control the probabilities of sampling the best feasible arm and the non-best-feasible set, similar in format to the approach of top-two sampling for BAI \citep*{qin2017improving,russo2020simple}. This new algorithm is called BFAI-TS. 

We perform a comprehensive analysis of the performance of BFAI-TS, including sample allocations of each arm and the rate of posterior convergence, which is the rate of the posterior probability of falsely selecting the best feasible arm converging to zero. This kind of theoretical characterization for BFAI algorithms has not be explored in the literature. We show that under BFAI-TS, each arm will be sufficiently sampled when the sample budget is large enough, which implies consistency of the algorithm (i.e., the estimated best arm will converge to the real best one). More importantly, we show that the rate of posterior convergence of BFAI-TS is on the exponential order and that BFAI-TS is asymptotically optimal, in the sense that this rate of the algorithm cannot be improved further.

In addition, the performance of BFAI-TS is demonstrated using numerical examples. We conducted experiments on five synthetic datasets and one realistic dataset to evaluate the performance of BFAI-TS and compare it against several benchmarks. The test results show that when the value of $\beta$ (for controlling the probabilities of sampling the best feasible arm and the non-best-feasible set) is properly set, BFAI-TS exhibits superior empirical performance relative to the compared algorithms.

\textbf{Related literature.} In this research, we employ TS \citep*{thompson1933likelihood} to design an algorithm to solve the BFAI problem. TS is a Bayesian algorithm for the multi-armed bandit (MAB) problem, aiming at minimizing the cumulative regret incurred during the sampling process. It is a simple sampling approach and has been extensively studied in various application domains, e.g., in Internet advertising \citep*{graepel2010web,agarwal2014laser}, recommendation systems \citep*{kawale2015efficient}, hyperparameter tuning \citep*{kandasamy2018parallelised}, etc. Due to its effectiveness, TS has also been extended to tackle a wide range of variant MAB problems, such as combinatorial bandits \citep*{sankararaman2018combinatorial}, contextual bandits \citep*{agrawal2013thompson} and
online problems \citep*{gopalan2014thompson}. 

Our algorithm also utilizes the idea of top-two sampling proposed for solving the BAI problem \citep*{russo2020simple}. This sampling method places emphasis on controlling the probabilities of sampling the best arm and the non-best set, which can be shown as a key step for optimizing the rate of posterior convergence of the sampling algorithms for BAI and its variant problems. In terms of the theoretical analysis, \citet*{qin2017improving} adopted a similar method for analyzing the rate of posterior convergence for BAI algorithms. In this research, we consider the BFAI problem, where the posterior convergence concerns not only the superiority of the arms on the objective measure but also the feasibility of the arms under constraint measures. This poses substantial challenges to the theoretical analysis of our study.

In the simulation community, the BFAI problem has been known as constrained ranking and selection (R\&S). \citet*{andradottir2010fully} tackled the problem of selecting the best feasible design with one constraint through a two-stage method. \citet*{lee2012approximate} solved constrained R\&S using the optimal computing budget allocation (OCBA) method \citep*{chen2000simulation,gao2017new,gao2017c,he2007opportunity,goodwin2024real}, by finding an analytical expression for the probability of correct selection and approximately optimizing it. \citet*{hunter2013optimal} generalized the work of \citet*{glynn2004large} from normal sampling distributions to general distributions. \citet*{pasupathy2014stochastically} developed a SCORE framework for efficient simulation budget allocation in large-scale constrained SO problems. \citet*{shi2019worst} further considered the problem when there is input uncertainty to the simulation model. These studies focus primarily on finding the sample allocations that approximately maximize the probabilities of correct selection for the best feasible arm, but lack theoretical characterization for the performance of the developed sampling algorithms, which we consider to be of importance for practitioners. In this research, we will fill this gap for our proposed BFAI-TS algorithm by investigating its rate of posterior convergence. We believe that the analysis presented in this paper can also be applied to characterize the theoretical properties of the aforementioned constrained R\&S algorithms.

\section{Problem Formulation}

Let $A=\{1,2,\ldots,k\}$ be the set of arms. When arm $i$ is pulled in round $t$, the agent gets a multidimensional noisy reward (sample) $\textbf{X}_{t,i}=[X_{t,i0},X_{t,i1},\ldots, X_{t,im}] \in \mathbb{R}^{m+1}$, where $X_{t,i0}$ is a sample of the objective measure for maximization and $X_{t,ij}$'s are samples of the constraint measures, $j=1,2,\ldots,m$. We assume that $X_{t,ij}$'s are independent across different rounds $t=1,2, \ldots, n$, arms $i=1,2,\ldots,k$ and measures $j=0,1,\ldots,m$, and follow normal distributions $\mathcal{N}(\mu_{ij},\sigma_{ij}^2)$ with known variances $\sigma_{ij}^2$. Means $\mu_{ij}$ are unknown and are learned by samples. Without loss of generality, we impose constraints $\mu_{ij} \leq \gamma_j$ on arms $i=1,2,\ldots,k$ and measures $j=1,2,\ldots,m$. If arm $i$ satisfies constraints $\mu_{ij}\leq\gamma_{j}$ for all $j=1,2,\ldots,m$, we call it a feasible arm; otherwise, it is infeasible. The goal of BFAI is to identify among feasible arms the one with the largest objective measure $\mu_{i0}$ with $n$ rounds of sampling, i.e., find $\argmax_{i\in A} \mu_{i0}$, s.t. $\mu_{ij}\leq\gamma_{j}$ for all $j=1,2,\ldots,m$. We assume that the best feasible arm is unique, and without loss of generality, let it be arm 1. In view that the arm set is finite, we also assume that no arms lie on the boundaries $\mu_{ij}=\gamma_j$ of the constraints to facilitate the analysis.

We consider the Bayesian framework and adopt normal distribution priors $\mathcal{N}(\mu_{1,ij},\sigma_{1,ij}^2)$. Suppose the algorithms pull arm $I_t$ in round $t$. With samples of the objective and constraint measures of the arms, we can calculate their posterior distributions. By conjugacy, the posterior distribution in round $t$ is also a normal distribution $\mathcal{N}(\mu_{t,ij}, \sigma_{t,ij}^{2})$, where the posterior mean and variance of arm $i$ and measure $j$ can be calculated by
\begin{equation*}
	\mu_{t+1,ij}=\left\{
	\begin{aligned}
		&\frac{\sigma_{t,ij}^{-2}\mu_{t,ij}+\sigma_{ij}^{-2}X_{t,ij}}{\sigma_{t,ij}^{-2}+\sigma_{ij}^{-2}}&
		{\mbox{if~}I_{t}=i},\\
		&\mu_{t,ij}&{\mbox{if~}I_{t}\neq i},
	\end{aligned}
	\right.
\end{equation*}
and
\begin{equation*}
	\sigma_{t+1,ij}^{2}=\left\{
	\begin{aligned}
		&\frac{1}{\sigma_{t,ij}^{-2}+\sigma_{ij}^{-2}}&
		{\mbox{if~}I_{t}=i},\\
		&\sigma_{t,ij}^{2}&{\mbox{if~}I_{t}\neq i}.
	\end{aligned}
	\right.
\end{equation*}
In this paper, we use non-informative priors, i.e., $\mu_{1,ij}=0$ and $\sigma_{1,ij}^{2}=\infty$ for all arms $i$ and measures $j$. Define $N_{t,i}\triangleq\sum_{l=1}^{t-1}\mathbf{1}\{I_{l}=i\}$ as the number of samples for arm $i$ before round $t$. In this case, we have 
\begin{equation*}
	\mu_{t,ij}=\frac{1}{N_{t,i}}\sum\limits_{l=1}^{t-1}\mathbf{1}\{I_{l}=i\}X_{l,I_{l}j}
\end{equation*}
and
\begin{equation*}
	\sigma_{t,ij}^{2}=\frac{1}{N_{t,i}}\sigma_{ij}^{2}.
\end{equation*}

Denote the posterior distribution over the vector of means of the objective and $m$ constraint measures by
\begin{equation*}
	\begin{split}
		\Pi_{t}=&\mathcal{N}(\mu_{t,10},\sigma_{t,10}^{2})\otimes\ldots\otimes\mathcal{N}(\mu_{t,1m},\sigma_{t,1m}^{2})
		\otimes\ldots
		\otimes\mathcal{N}(\mu_{t,k0},\sigma_{t,k0}^{2})\otimes\ldots\otimes\mathcal{N}(\mu_{t,km},\sigma_{t,km}^{2}).
	\end{split}
\end{equation*}

Suppose $\theta=[\theta_{10},\ldots,\theta_{1m},\ldots,\theta_{k0},\ldots,\theta_{km}]$ is sampled from distribution $\Pi_t$.
The (posterior) probability that arm $i$ is estimated as the best feasible arm is 

\begin{equation*}
	\begin{split}
		P_{t,i}\triangleq &\mathbb{P}_{\theta\sim\Pi_{t}}\Bigg(\bigcap_{i'\neq i}\Big((\theta_{i0}<\theta_{i^{'}0})\cap
		\bigcap_{j=1}^{m}(\theta_{i^{'}j}\leq\gamma_{j})\Big)^{c}
		\cap\bigcap_{j=1}^{m}(\theta_{ij}\leq\gamma_{j})\Bigg),
	\end{split}
\end{equation*}
where $Y^{c}$ refers to the complement of event $Y$. The event $\Big((\theta_{i0}<\theta_{i^{'}0})\cap\bigcap_{j=1}^{m}(\theta_{i^{'}j}\leq\gamma_{j})\Big)^{c}$ can rule out the possibility of arm $i'$ being chosen as the best feasible arm, while the event $\bigcap_{j=1}^{m}(\theta_{ij}\leq\gamma_{j})$ can capture the feasibility of arm $i$. 
\begin{algorithm}[tb]
	\caption{BFAI-TS Algorithm}
	\label{BFAI-TS}
	\begin{algorithmic}
		\REQUIRE $k\geq2$, $\beta\in(0,1)$, $n$
		\STATE Collect $n_{0}$ samples for each arm $i$.
		\WHILE{$t\leq n$}
		\STATE Sample $\theta\sim\Pi_{t}$
		\STATE Get the feasible set $F\triangleq\{i:\theta_{ij}\leq\gamma_{j}	\mbox{~for~} j =1,2,\ldots,m\}$
		\IF{$F\neq\emptyset$}
		\STATE Set $I_{t}^{(1)}\leftarrow\argmax\theta_{i0}$ for $i\in F$
		\ELSE
		\STATE Choose $I_{t}^{(1)}$ uniformly from $\{1,2,\ldots,k\}$
		\ENDIF
		\STATE Sample $B\sim$ Bernoulli$(\beta)$
		\IF{$B=1$}
		\STATE Play $I_{t}^{(1)}$
		\ELSE 
		\REPEAT
		\STATE Sample $\theta\sim\Pi_{t}$
		\STATE Get the feasible set $F\triangleq\{i:\theta_{ij}\leq\gamma_{j}	\mbox{~for~} j =1,2,\ldots,m\}$
		\IF{$F\neq\emptyset$}
		\STATE Set $I_{t}^{(2)}\leftarrow\argmax\theta_{i0}$ for $i\in F$
		\ELSE
		\STATE Choose $I_{t}^{(2)}$ uniformly from $\{1,2,\ldots,k\}$
		\ENDIF
		\UNTIL{$I_{t}^{(2)}\neq I_{t}^{(1)}$}
		\STATE Play $I_{t}^{(2)}$
		\ENDIF
		\STATE Update posterior $\Pi_{t+1}$
		\ENDWHILE
		\ENSURE $I^{*}$
	\end{algorithmic}
\end{algorithm}

\section{BFAI-TS Algorithm}

TS is a popular sampling method for bandit problems. It is built on the simple idea of pulling an arm according to its posterior probability of being the best arm. To adapt this idea to BFAI, it is natural to iteratively pull the arm according to the posterior probability that it is the best feasible one. However, this natural extension does not work well. We notice that it is not asymptotically optimal. Although the sampling ratios of any pair of arms in the non-best-feasible set are optimal by this method, the algorithm tends to allocate too many samples to the best feasible arm in the long run. To fix it, we introduce a parameter $\beta\in (0,1)$ to control the probabilities of sampling the best feasible arm and the non-best-feasible set, similar in format to the approach of top-two sampling for BAI.

In each round, samples of each arm are generated from their posterior distributions. The agent identifies the arm $I_{t}^{(1)}$ with the best feasible posterior sample. If none of the samples from the arms are feasible, the agent uniformly chooses an arm from $\{1,2,...,k\}$ as $I_{t}^{(1)}$. With probability $\beta$, the agent pulls arm $I_{t}^{(1)}$. Otherwise, the agent pulls an arm in the non-best-feasible set according to their posterior probability of being the best feasible. In other words, with probability $1-\beta$, the agent identifies arm $I_{t}^{(2)}$ and pulls it, where $I_{t}^{(2)}$ is the arm with the best feasible posterior sample from the set $\{1,2,...,k\} \backslash\{I_{t}^{(1)}\}$. Again, if none of the samples from the arms are feasible, the agent uniformly chooses an arm from $\{1,2,...,k\}$ that is not $I_{t}^{(1)}$ as $I_{t}^{(2)}$. Let $\mathcal{F}_{t}$ denote the sigma algebra generated by $(I_{1}, \textbf{X}_{t,I_{1}}, \ldots, I_{t}, \textbf{X}_{t,I_{t}} )$. For all $i$ and $t\in \mathbb{N}$, define $\phi_{t,i} \triangleq\mathbb{P}(I_{t}=i|\mathcal{F}_{t-1})$ and $\bar{\phi}_{t,i}\triangleq\frac{\sum_{l=2}^{t}\phi_{l,i}}{t}$. Then, the probability of pulling arm $i$ in round $t$ is
\begin{equation}
	\begin{split}
		\phi_{t,i}=&\frac{c_{t}}{k}+(1-\beta)P_{t,i}\sum_{i'\neq i}\Big(\frac{P_{t,i'}}{1-P_{t,i'}}(1-c_{t})+\frac{c_{t}}{k-1}\Big)
		+P_{t,i}\beta(1-c_{t}),
	\end{split}
\end{equation}
where $c_{t}$ is the probability that samples from all the arms are infeasible in round $t$. We can see that as $P_{t,1}\rightarrow1$, $\phi_{t,1}\rightarrow\beta $, $c_{t}\rightarrow 0$ and for each arm $i\neq 1$, $\frac{P_{t,i}}{1-P_{t,1}}\rightarrow\frac{\phi_{t,i}}{1-\phi_{t,1}}$.

\section{Theoretical Results}
In this section, the performance of the BFAI-TS Algorithm will be theoretically characterized. We will show that its posterior probability of false selection is on the exponential order and is governed by a dominant term, which is referred to as the rate of posterior convergence $\Gamma$, i.e., $\Gamma=\lim\limits_{n\rightarrow\infty}-\frac{1}{n}\log (1-P_{n,1})$. We will establish the asymptotic optimality of its sample allocations and rate of posterior convergence.

We denote the set of feasible arms as $\mathcal{F}\triangleq\{i:\mu_{ij}\leq\gamma_j, \forall j\in \{1,2,...,m\}\}$. To ease the discussion, we partition the $k$ arms into the following four mutually exclusive subsets.
\begin{enumerate}
	\item [$I^{*}$] the best feasible arm, i.e., $I^{*}\triangleq\argmax_{i\in \mathcal{F}}\mu_{i0}$;
	\item[$\mathcal{F}_{w}$] the set of feasible but suboptimal arms, i.e., $\mathcal{F}_{w}\triangleq\{i: i\in\mathcal{F}\mbox{ and }i\neq I^{*}\}$;
	\item[$\mathcal{I}_{b}$] the set of infeasible arms with objective performance no worse than $I^*$,
	i.e., $\mathcal{I}_{b}\triangleq\{i: \mu_{I^{*}0}\leq \mu_{i0} ~\mbox{and}~\exists j\in\{1,2,\ldots,m\} \mbox{~such~ that~} \mu_{ij}>\gamma_{j}\}$;
	\item[$\mathcal{I}_{w}$] the set of infeasible arms with objective performance worse than $I^{*}$, i.e., $\mathcal{I}_{w}\triangleq\{i: \mu_{I^{*}0}> \mu_{i0} ~\mbox{and}~\exists j\in\{1,2,\ldots,m\} \mbox{~such~ that~}~\mu_{ij}>\gamma_{j}\}$.
\end{enumerate}

For arm $i$, we also classify the $m$ constraints into two subsets for which the arm satisfies and violates.
Let $\mathcal{M}_{F}^{i}$ be the set of constraints satisfied by arm $i$, i.e., $\{j: \mu_{ij}\leq \gamma_{j} \mbox{~for~} j \in\{1,2,\ldots,m\}\}$ and $\mathcal{M}_{I}^{i}$ be the set of constraints violated by arm $i$, i.e., $\{j: \mu_{ij}> \gamma_{j} \mbox{~for~} j \in\{1,2,\ldots,m\}\}$.

%
%

Three definitions will be used in our analysis. First, according to the construction of the algorithms, it is obvious that the sampling rate of arm 1 converges to $\beta$ with $\lim_{n \rightarrow \infty} N_{n,1}/n=\beta$. Denote the optimal sampling rates of the remaining $k-1$ arms by the vector ($\alpha_2^\beta,...,\alpha_k^\beta$) with $\sum_{i=2}^k \alpha_i^\beta=1-\beta$. Here, by optimal sampling rates, 
we mean that the sampling rates can lead the algorithm to achieve the fastest possible rate of posterior convergence $\Gamma$ for the probability of false selection $1-P_{n,1}$ among all algorithms allocating $\beta$ proportion of the total samples to the best feasible arm.
As will be shown in the proof of Theorem \ref{thm1}, the optimal sampling rates ($\alpha_{2}^{\beta},\ldots,\alpha_{k}^{\beta}$) satisfy the following optimality condition
\begin{equation}\label{optimal condition}
	\sum_{i=2}^{k}\alpha_{i}^{\beta}=1-\beta, \text{ and } \mathcal{R}_i=\mathcal{R}_{i'} \text{ for any }i\neq i^{'}\neq 1,
\end{equation}
where $\mathcal{R}_i=\frac{(\mu_{i0}-\mu_{10})^{2}}{(\sigma_{i0}^{2}/\alpha_{i}^{\beta}+\sigma_{10}^{2}/\beta)}\mathbf{1}\{i\in\mathcal{F}_{w}\cup \mathcal{I}_{w}\}	+\alpha_{i}^{\beta}\sum\limits_{j\in\mathcal{M}_{I}^{i}}\frac{(\mu_{ij}-\gamma_{j})^{2}}{\sigma_{ij}^{2}}\mathbf{1}\{i\in\mathcal{I}_{b}\cup \mathcal{I}_{w}\}$.

Second, given $\beta\in(0,1)$ and the optimal sampling rates, define $\Gamma_\beta$ as
\begin{equation}
	\begin{split}
		\Gamma_\beta=&\min_{i\neq 1}\bigg(\frac{(\mu_{i0}-\mu_{10})^{2}}{2(\sigma_{i0}^{2}/\alpha_{i}^{\beta}+\sigma_{10}^{2}/\beta)}\mathbf{1}\{i\in\mathcal{F}_{w}\cup \mathcal{I}_{w}\}+\alpha_{i}^{\beta}\sum\limits_{j\in\mathcal{M}_{I}^{i}}\frac{(\mu_{ij}-\gamma_{j})^{2}}{2\sigma_{ij}^{2}}\mathbf{1}\{i\in\mathcal{I}_{b}\cup \mathcal{I}_{w}\},\\
		&\min\limits_{j\in\mathcal{M}_{F}^{1}}\beta\frac{(\mu_{1j}-\gamma_{j})^{2}}{2\sigma_{1j}^{2}}\bigg).
	\end{split}
\end{equation}
In particular, when $\beta$ is set to the optimal value $\beta^*$, according to (\ref{optimal condition}), $\Gamma_{\beta^{*}}=\frac{(\mu_{i0}-\mu_{10})^{2}}{2(\sigma_{i0}^{2}/\alpha_{i}^{*}+\sigma_{10}^{2}/\beta^{*})}\mathbf{1}\{i\in\mathcal{F}_{w}\cup \mathcal{I}_{w}\}+\alpha_{i}^{*}\sum\limits_{j\in\mathcal{M}_{I}^{i}}\frac{(\mu_{ij}-\gamma_{j})^{2}}{2\sigma_{ij}^{2}}\mathbf{1}\{i\in\mathcal{I}_{b}\cup \mathcal{I}_{w}\}$. It will be shown in Theorem \ref{thm2} that $\Gamma_\beta$ and $\Gamma_{\beta^{*}}$ characterize the optimal rate of posterior convergence for $1-P_{n,1}$.

Third, we introduce a state in which the posterior means and sampling rates are accurate enough compared to the true means and optimal sampling rates. Given $\beta\in(0,1)$ and $\epsilon>0$, let
\begin{equation}\label{Nbeta}
	\begin{split}
		N_{\beta}^{\epsilon}&\triangleq\inf\{t\in\mathbb{N}: |\mu_{n,ij}-\mu_{ij}|\leq\epsilon\mbox{~and~}
		|N_{n,i}/n-\alpha_{i}^{\beta}|\leq\epsilon, \forall i\in A\mbox{~and~}n\geq t\}.
	\end{split}
\end{equation}
It's obvious that if $N_{n,i}/n\rightarrow \alpha_{i}^{\beta}$ with probability 1, $\mathbb{P}(N_{\beta}^{\epsilon}<\infty)=1$ for any $\epsilon>0$.

\begin{theorem}\label{thm1}
	For the BFAI-TS Algorithm, $\mathbb{E}[N_{\beta}^{\epsilon}]<\infty$ for any $\epsilon>0$. The sample allocations of the algorithm is asymptotically optimal in the sense that
	\begin{equation}
		\lim_{n\rightarrow \infty}\frac{N_{n,i}}{n}\xrightarrow{p}\alpha_{i}^{\beta}\quad\forall i\in A,
	\end{equation}
\end{theorem}
where $\xrightarrow{p}$ means the property holds with probability $1$.

In other words, on average, it takes a finite number of rounds for the algorithm to be ``accurate enough".

Next, we characterize the rate of posterior convergence of the BFAI-TS algorithm and establish the optimality of it. Note that the rate depends on parameter $\beta$. Below we consider this rate under two cases when $\beta$ takes any value in $(0,1)$ and when $\beta$ is set to the optimal value $\beta^*$ (value that maximizes the rate of posterior convergence of BFAI-TS).

\begin{theorem}\label{thm2}	
	Following properties hold with probability 1:\\
	1. For any $\beta\in(0,1)$, $\Gamma_\beta$ shows the fastest rate of posterior convergence that any algorithm allocating $\beta$ proportion of the total samples to the best feasible arm can possibly achieve
	\begin{equation}
		\limsup\limits_{n\rightarrow\infty}-\frac{1}{n}\log (1-P_{n,1})\leq\Gamma_{\beta}
	\end{equation}
	and the BFAI-TS Algorithm achieves this rate with
	\begin{equation}
		\lim\limits_{n\rightarrow\infty}-\frac{1}{n}\log (1-P_{n,1})=\Gamma_{\beta}.
	\end{equation}		
	2. The term $\Gamma_{\beta^*}$ shows the fastest rate of posterior convergence that any BFAI algorithm can possibly achieve
	\begin{equation}
		\limsup\limits_{n\rightarrow\infty}-\frac{1}{n}\log (1-P_{n,1})\leq\Gamma_{\beta^{*}}
	\end{equation}
	and when the $\beta$ of the BFAI-TS Algorithm is set to $\beta^*$, the algorithm achieves the optimal rate with
	\begin{equation}
		\lim\limits_{n\rightarrow\infty}-\frac{1}{n}\log (1-P_{n,1})=\Gamma_{\beta^{*}}.
	\end{equation}		
\end{theorem}

The first part of Theorem \ref{thm2} establishes the asymptotic optimality of the BFAI-TS algorithm in the sense that, under any given $\beta\in (0,1)$, the rate of posterior convergence for the probabilities of false selection of the algorithm is the fastest possible. Other algorithms cannot perform better than this rate. The second part of the theorem shows the same result for the BFAI-TS algorithm when $\beta$ is set to the optimal value $\beta^*$. The sampling decisions of any algorithm in each round are based on prior samples, which are inherently random.

\section{Numerical Results}

\begin{table*}
	\caption{Probabilities of false selection for the tested algorithms.}
	\label{table 1}
	\centering
	\resizebox{\linewidth}{!}{
		\begin{tabular}{|c|c|c|c|c|c|c|c|c|c|c|c|c|c|c|c|c|c|c|}
			\hline
			Experiments&\multicolumn{3}{|c|}{Experiment 1}&\multicolumn{3}{|c|}{Experiment 2}&\multicolumn{3}{|c|}{Experiment 3}&\multicolumn{3}{|c|}{Experiment 4}&\multicolumn{3}{|c|}{Experiment 5}&\multicolumn{3}{|c|}{Dose-finding}\\
			\cline{1-19}
			\diagbox[dir=SE]{Algorithms}{Sample size}&2100&2500&3400&1000&2500&3500&2000&2400&3700&2600&3300&3700&200&400&800&3500&6000&8000\\		
			\hline
			\cline{1-19}
			BFAI-TS-1&0.20&0.17&0.13&0.38&0.14&0.10&0.24&0.19&0.12&0.13&0.06&0.04&0.37&0.25&0.22&0.18&0.09&0.06\\
			\cline{1-19}
			OCBA-CO&0.13&0.10&0.04&0.25&0.14&0.05&0.14&0.11&0.06&0.09&0.06&0.05&0.31&0.24&0.20&0.20&0.12&0.10\\
			\cline{1-19}
			Uniform&0.61&0.51&0.46&0.71&0.59&0.50&0.46&0.48&0.39&0.54&0.49&0.51&0.37&0.41&0.34&0.27&0.19&0.13\\
			\cline{1-19}
			TF-LUCB&0.22&0.11&0.05&0.51&0.15&0.06&0.37&0.24&0.11&0.26&0.13&0.09&0.27&0.19&0.21&0.16&0.10&0.09\\
			\cline{1-19}
			MD-UCBE&0.87&0.87&0.87&0.82&0.85&0.83&0.79&0.79&0.77&0.79&0.77&0.77&0.50&0.50&0.50&0.21&0.19&0.19\\
			\cline{1-19}
			BFAI-TS ($\beta=0.5$)&0.11&0.06&0.02&0.24&0.09&0.03&0.12&0.07&0.05&0.04&0.02&0.02&0.25&0.18&0.15&0.13&0.07&0.05\\
			\cline{1-19}
			BFAI-TS ($\beta=\beta^{*}$)&0.07&0.02&0.01&0.19&0.02&0.01&0.09&0.05&0.01&0.02&0.01&0.00&0.11&0.03&0.00&0.11&0.05&0.01\\
			\hline
		\end{tabular}
	}
\end{table*}

In this section, we show the empirical performances of the BFAI-TS Algorithm on synthetic and real-world datasets. For comparison, we will test the uniform allocation, MD-UCBE Algorithm \citep{katz2018feasible}, TF-LUCB Algorithm \citep{katz2019top} and OCBA-CO Algorithm \citep{lee2012approximate}. 

The uniform allocation is a simple benchmark which pulls each arm with the same probability $1/k$ in each round. The MD-UCBE Algorithm aligns with our algorithms in the fixed-budget setting but it can only identify feasible arms. The TF-LUCB Algorithm aligns with our algorithms in the objective of identifying the best (top) feasible arm(s) but it falls in the fixed-confidence setting. The OCBA-CO Algorithm calculates the estimated optimal allocation rule and allocates the budget by it in each round.

We also test a special case of our algorithm called BFAI-TS-1. It is BFAI-TS with $\beta=1$. BFAI-TS-1 can be treated as the results of direct applications of the ideas of TS to BFAI without the use of the top-two framework and $\beta$. The proposed BFAI-TS is tested when $\beta =0.5$ and when $\beta$ takes the optimal value $\beta^*$. In practice, a simple and robust empirical setting is to let $\beta=0.5$. If we want to adopt a more effective setting of $\beta$, we note that as $\beta$ increases, the value of $\min_{i\neq I^{*}}\mathcal{R}_{i}$ will first increase and then decrease, and the optimal value $\beta^{*}$ of $\beta$ is the maximizer of $\min_{i\neq I^{*}}\mathcal{R}_{i}$. Using this fact, we can estimate $\mathcal{R}_{i}$ using the posterior mean $\mu_{n,i}$ of arm $i$ and adaptively change $\beta$ to increase $\min_{i\neq I^{*}}\mathcal{R}_{i}$ in the algorithms, such that $\beta$ gets closer to $\beta^{*}$.

\textbf{Synthetic Datasets}. Experiments 1-4 have $50$ arms, corresponding to $x=1,2,\ldots,50$, and the best feasible arm is always $x= 26$. Experiment 5 has $10$ arms, corresponding to $x=1,2,\ldots,10$, and the best feasible arm is $x=10$. All the tested algorithms collect six samples of each arm to generate priors. Samples of the objective and constraint measures are corrupted by normal noises. Let the constraint limits $\gamma_{j}=0$ for $j=1,2,\ldots,m$, because all constraints $\mu_{ij}\leq\gamma_{j}$ can be transformed into $\mu_{ij}-\gamma_{j}\leq 0$. The performances of the algorithms are obtained based on the average of 100 macro-replications.

Experiment 1 (One constraint) We consider the objective function 
$y_1(x)=0.08(1-x)$ and constraint $y_2(x)\leq0$, where
\begin{equation*}
	y_2(x)=\left\{
	\begin{aligned}
		&0.08(26-x)&
		{\mbox{if~}1\leq x\leq 25},\\
		&0.08(-x+25)&{\mbox{if~}26\leq x\leq 50}.
	\end{aligned}
	\right.
\end{equation*}\\
Noises of $y_1(x)$ and $y_2(x)$ follow the same normal distribution $\mathcal{N}(0,0.49)$. By calculation, $\beta^{*}\approx 0.3218$.

Experiment 2 (Four constraints) We retain the objective function and constraint of Experiment 1, and add three constraints: constraint $y_3(x)\leq0$, where 
\begin{equation*}
	y_3(x)=\left\{
	\begin{aligned}
		&0.01(21-x)^{2}&
		{\mbox{if~}1\leq x\leq 20},\\
		&-0.01(-x+20)^{2}&{\mbox{if~}21\leq x\leq 50};
	\end{aligned}
	\right.
\end{equation*}\\
constraint $y_4(x)\leq0$, where 
\begin{equation*}
	y_4(x)=\left\{
	\begin{aligned}
		&-0.1(41-x)&
		{\mbox{if~}1\leq x\leq 40},\\
		&-0.1(-x+40)&{\mbox{if~}41\leq x\leq 50};
	\end{aligned}
	\right.
\end{equation*}\\
and constraint $y_5(x)\leq0$, where 
\begin{equation*}
	y_5(x)=\left\{
	\begin{aligned}
		&-0.03(41-x)&
		{\mbox{if~}1\leq x\leq 40},\\
		&-0.03(-x+40)&{\mbox{if~}41\leq x\leq 50}.
	\end{aligned}
	\right.
\end{equation*}\\
Noises of $y_1(x)$-$y_5(x)$ follow the same normal distribution $\mathcal{N}(0,0.49)$. By calculation, $\beta^{*}\approx 0.2449$.

Experiment 3 (Four constraints) We retain the objective function and all constraints of Experiment 2. Distributions of the noises of $y_1(x)$-$y_5(x)$ are changed to $\mathcal{N}(0,0.36)$, $\mathcal{N}(0,0.81)$, $\mathcal{N}(0,0.64)$, $\mathcal{N}(0,0.49)$ and $\mathcal{N}(0,1)$ respectively. By calculation, $\beta^{*}\approx 0.2709$.

Experiment 4 (Four constraints) We retain the objective function $y_1(x)$ and constraints $y_2(x)\leq0$, $y_4(x)\leq0$, and $y_5(x)\leq0$ of Experiment 3, and replace the constraint $y_3(x)\leq0$ by $y'_3(x)\leq0$, where 
\begin{equation*}
	y'_3(x)=\left\{
	\begin{aligned}
		&0.01(21-x)^{2}+0.1&
		{\mbox{if~}1\leq x\leq 20},\\
		&-0.01(-x+20)^{2}-0.1&{\mbox{if~}21\leq x\leq 50}.
	\end{aligned}
	\right.
\end{equation*}\\
This change reduces the difficulty of correctly identifying the feasibility of this constraint, and thus favors the compared MD-UCBE and TF-LUCB Algorithms which focus more on feasibility detection. By calculation, $\beta^{*}\approx 0.2615$.

Experiment 5 (One constraint) The means in objective measure of the ten arms are -1.8455, 0.2556, -1.7275, 0.0219, -1.0574, 1.7303, 1.6237, 1.8268, -1.6826 and 1.8150, and the means in constraint measure of the ten arms are -1.8441, -0.0028, 0.4682, -1.0172, 0.6831, -1.5495, 0.1442, 1.2425, 1.3175 and -0.6513, all of which are uniformly generated in $(-2,2)$. If the mean of constraint measure is less than zero, the arm is feasible. Samples of the arms are corrupted by normal noises $\mathcal{N}(0,1)$. By calculation, $\beta^{*}\approx 0.4831$.

\textbf{Dose-Finding Problem}. In clinical trials, it is important to find the most effective dosage of a drug, while keeping the probability of the drug causing an adverse effect below some safety threshold. In this test, we use the data in \citet*{genovese2013efficacy} (see ARCR$20$ in week $16$). It studies the drug secukinumab for treating rheumatoid arthritis. There are four dosage levels, $25$mg, $75$mg, $150$mg, and $300$mg, and a placebo, which are treated as five arms. We design a simulation model based on the dataset. Each arm is associated with two performance measures: the probability of the drug being effective and the probability of the drug causing infections. The mean performances of the five arms are $\boldsymbol{\mu_{1}}=(0.34,0.259)$, $\boldsymbol{\mu_{2}}=(0.469,0.184)$, $\boldsymbol{\mu_{3}}=(0.465,0.209)$, $\boldsymbol{\mu_{4}}=(0.537,0.293)$ and $\boldsymbol{\mu_{5}}=(0.36,0.16)$. We assume that a drug is acceptable if the probability of infections is below $0.25$, and observations of each arm are corrupted by normal noises with means 0 and variances $0.01$. The best arm is arm $2$ (dosage level 75mg). By calculation, $\beta^{*}\approx 0.4986$.

Table 1 shows the probabilities of false selection for the best feasible arm under the compared algorithms and different sample sizes. The proposed BFAI-TS outperforms the other algorithms, especially when the sample size is small. The setting of $\beta=\beta^*$ is better than $\beta=0.5$. If we abandon the top-two framework and let $\beta=1$, the performances of the algorithms deteriorate a lot. It shows the importance of introducing $\beta$ to control the proportion of samples allocated to the best feasible arm. MD-UCBE, TF-LUCB and OCBA-CO are inferior to the proposed BFAI-TS algorithm under proper values of $\beta$. This is not surprising. MD-UCBE and TF-LUCB do not directly fall in the fixed-budget setting of the best feasible arm identification. MD-UCBE is only concerned with the feasibility of the arms and ignores their objective performances. TF-LUCB is a fixed-confidence algorithm that focuses on making guarantees on the probability of false selection instead of minimizing it. The OCBA-CO algorithm allocates samples following the estimated optimal allocation rule, which aims to maximize the approximation of the probability of the correct selection instead of the real probability of the correct selection. The performance of the uniform allocation is stable, and is in general better than MD-UCBE.

\begin{figure*}[htbp]
	\centering
	\subfigure[]{
		\begin{minipage}[t]{0.5\linewidth}
			\centering
			\includegraphics[width=2.8in]{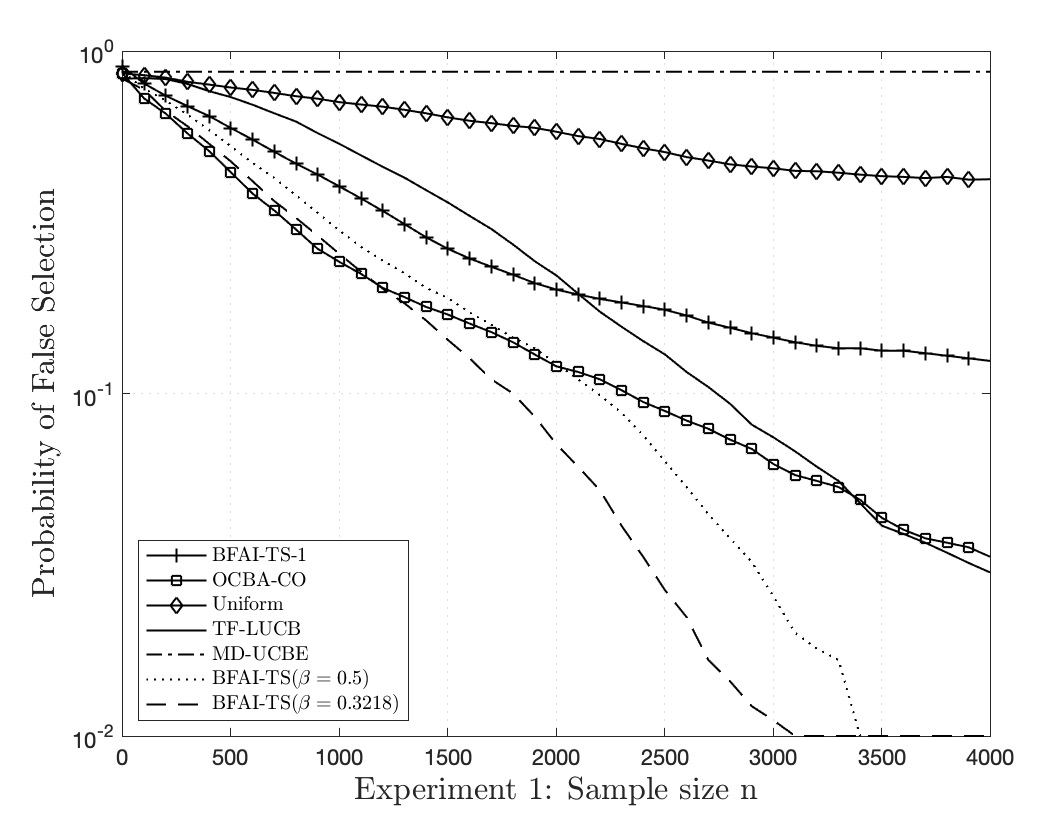}
		\end{minipage}%
	}%
	\subfigure[]{
		\begin{minipage}[t]{0.5\linewidth}
			\centering
			\includegraphics[width=2.8in]{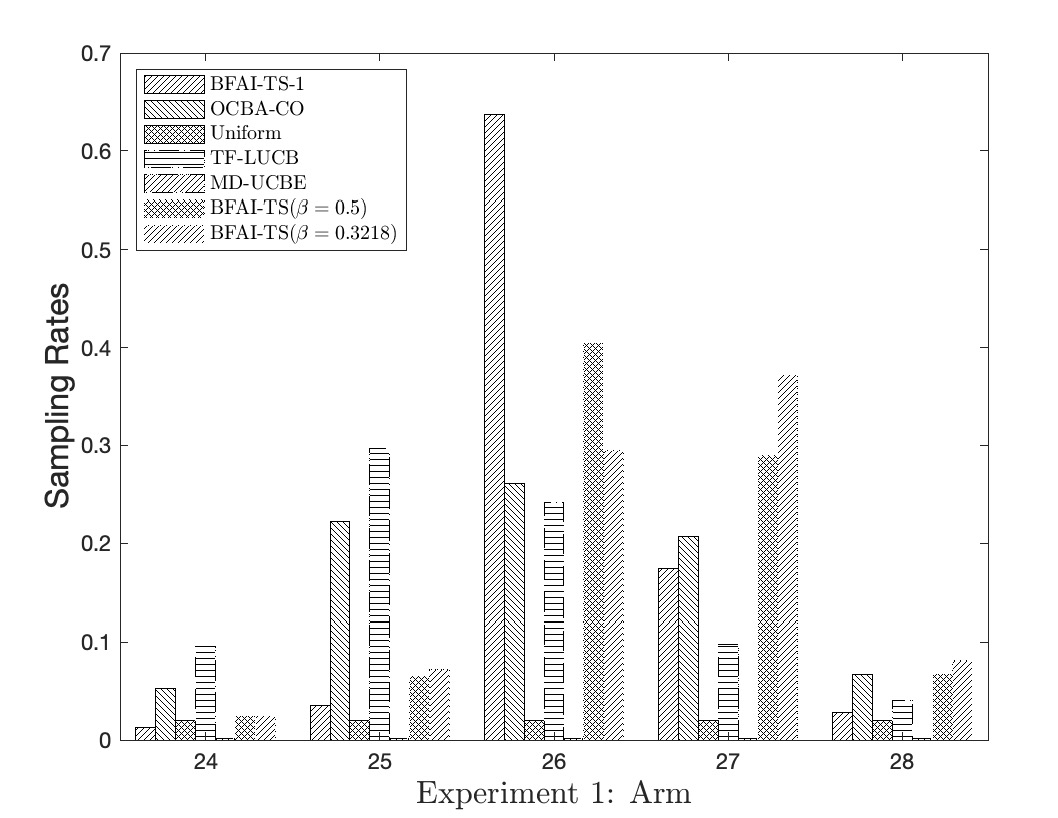}
		\end{minipage}%
	}%
	\centering
	\caption{PFS and their sampling rates on selected arms (Experiment 1)}
	
	\subfigure[]{
		\begin{minipage}[t]{0.5\linewidth}
			\centering
			\includegraphics[width=2.8in]{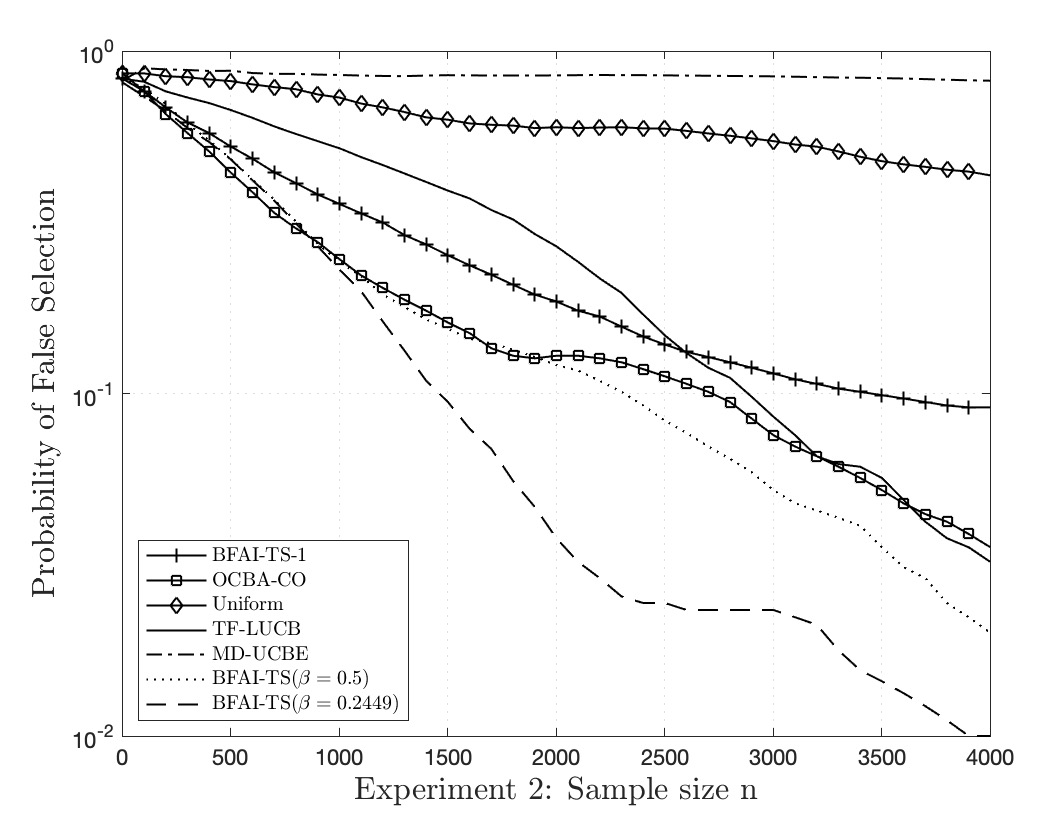}
		\end{minipage}%
	}%
	\subfigure[]{
		\begin{minipage}[t]{0.5\linewidth}
			\centering
			\includegraphics[width=2.8in]{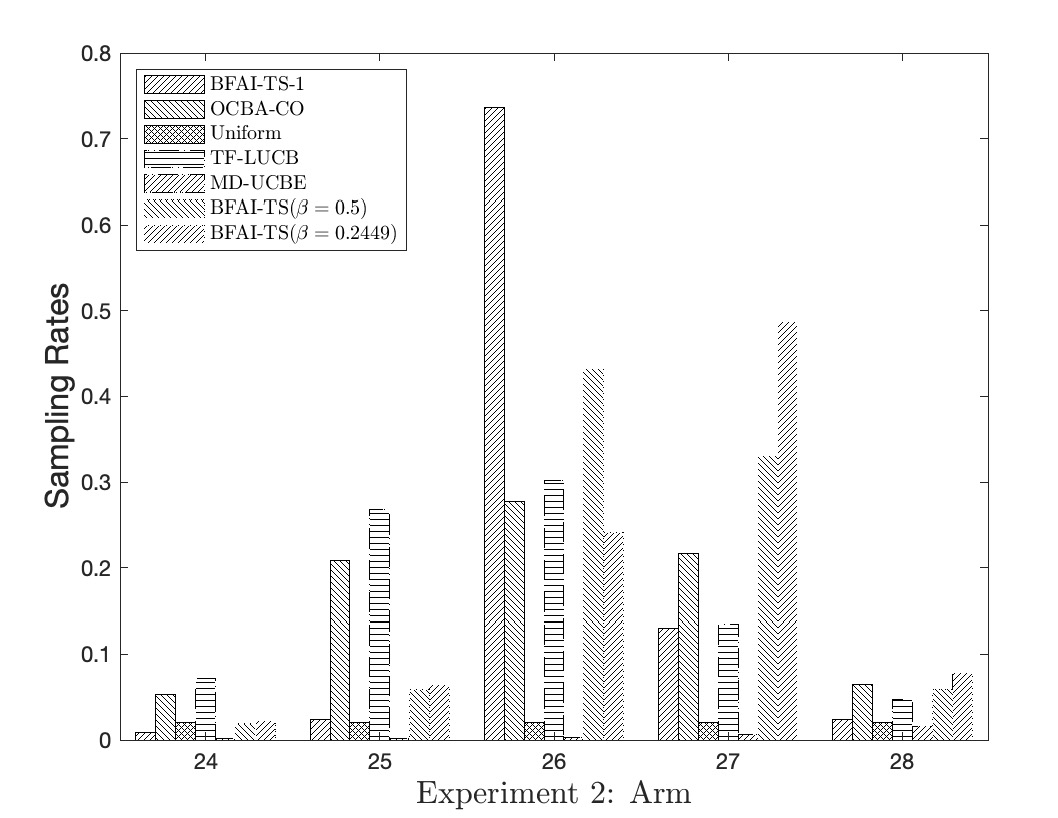}
		\end{minipage}%
	}%
	\centering
	\caption{PFS and their sampling rates on selected arms (Experiment 2)}
	
	\subfigure[]{
		\begin{minipage}[t]{0.5\linewidth}
			\centering
			\includegraphics[width=2.8in]{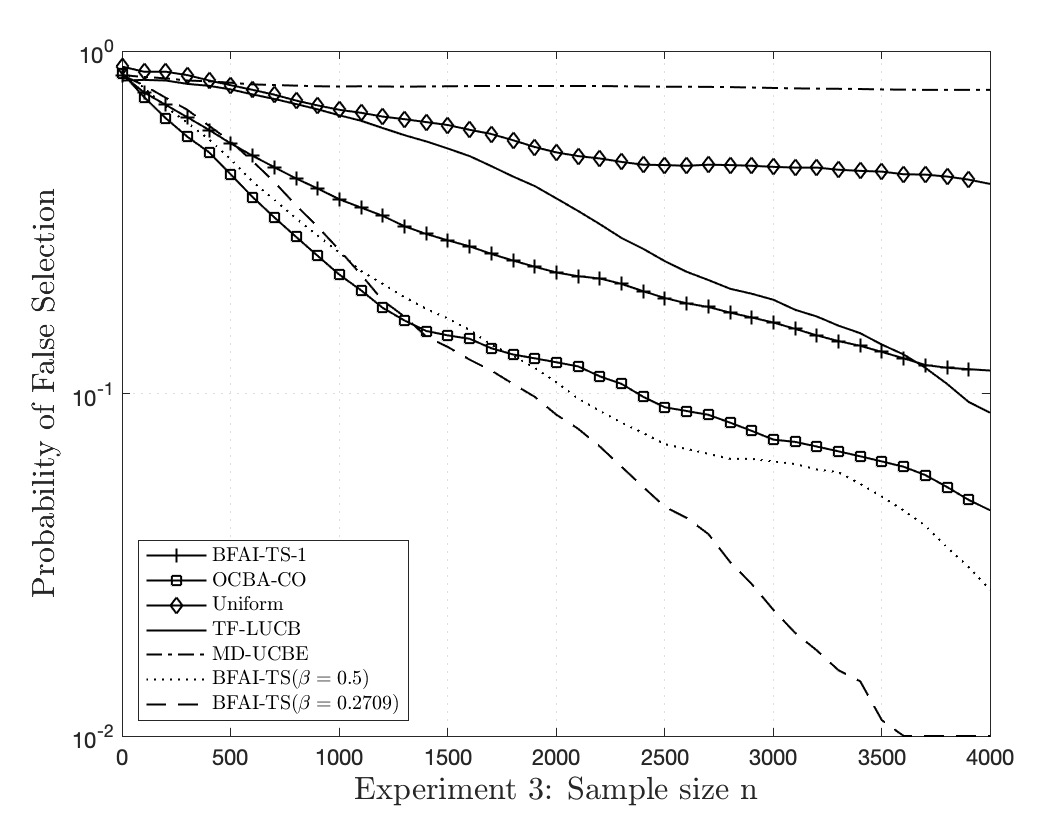}
		\end{minipage}%
	}%
	\subfigure[]{
		\begin{minipage}[t]{0.5\linewidth}
			\centering
			\includegraphics[width=2.8in]{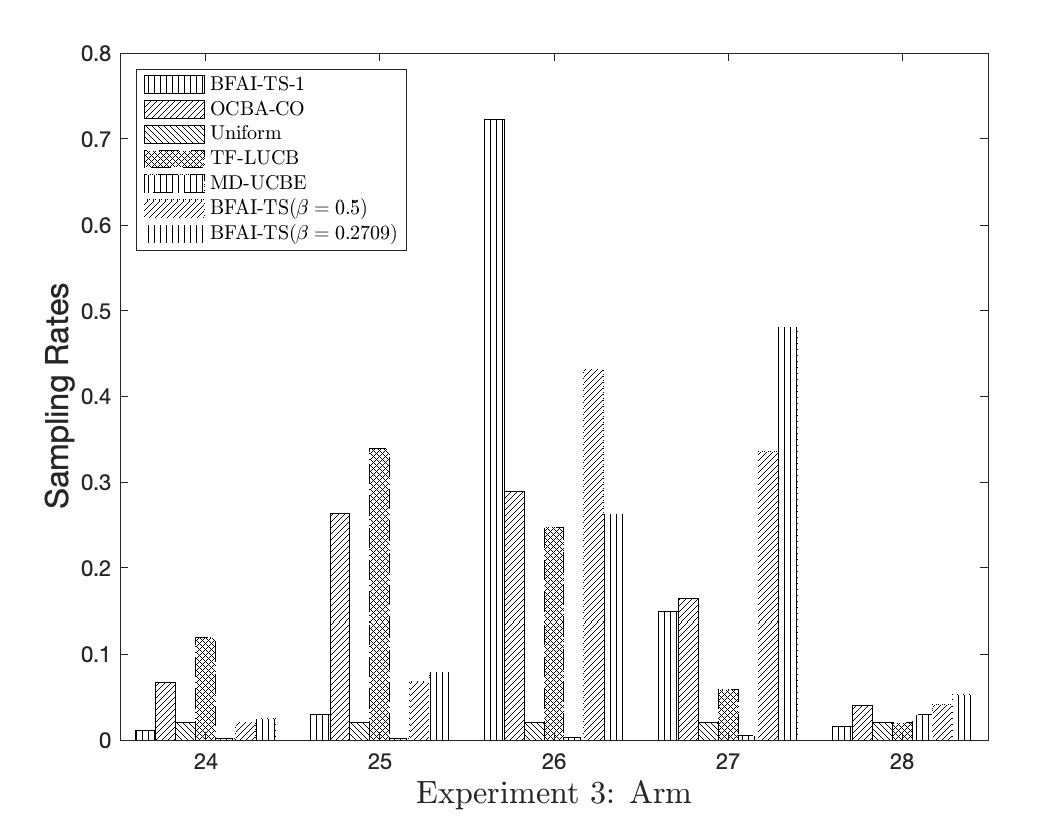}
		\end{minipage}%
	}%
	\centering
	\caption{PFS and their sampling rates on selected arms (Experiment 3)}
\end{figure*}	
\begin{figure*}[htbp]
	\centering	
	\subfigure[]{
		\begin{minipage}[t]{0.5\linewidth}
			\centering
			\includegraphics[width=2.8in]{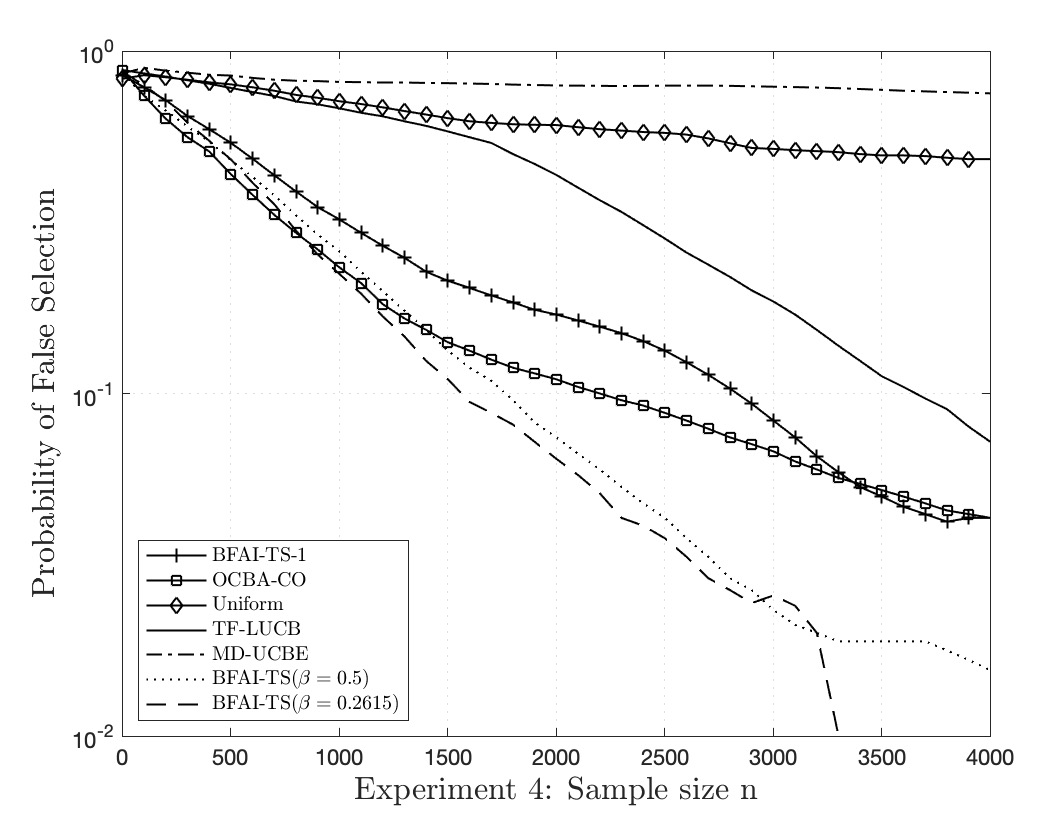}
		\end{minipage}%
	}%
	\subfigure[]{
		\begin{minipage}[t]{0.5\linewidth}
			\centering
			\includegraphics[width=2.8in]{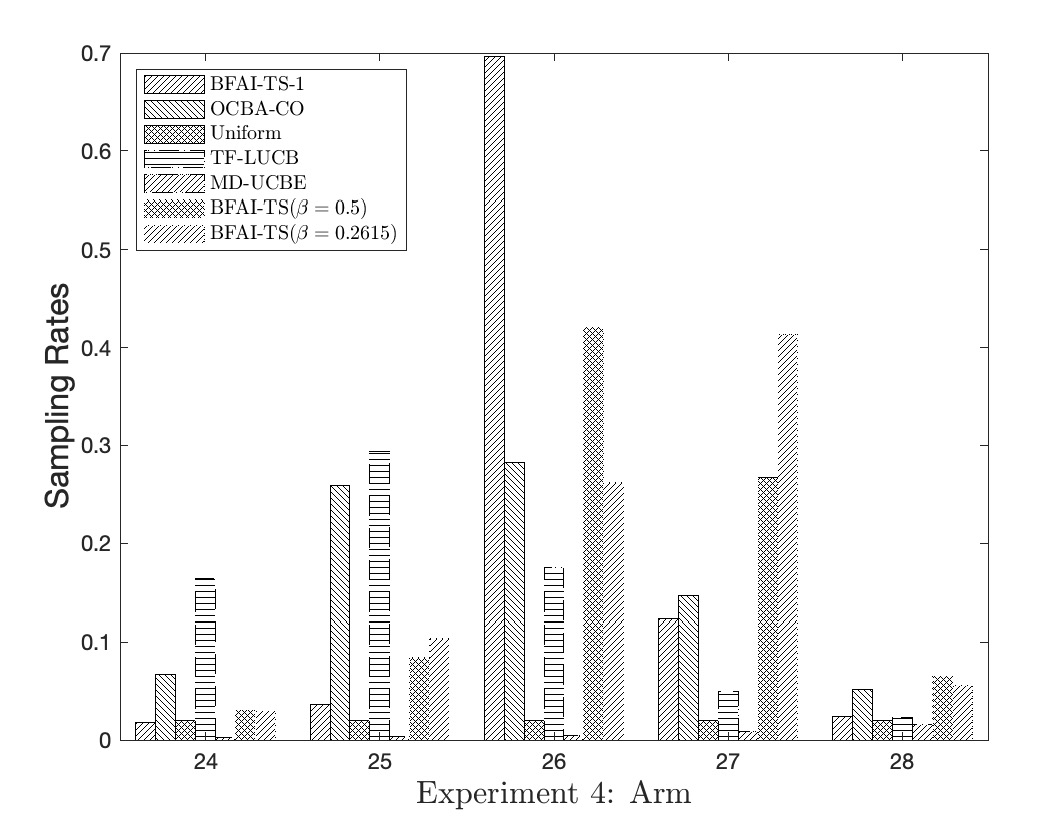}
		\end{minipage}%
	}%
	\centering
	\caption{PFS and their sampling rates on selected arms (Experiment 4)}
	
	\subfigure[]{
		\begin{minipage}[t]{0.5\linewidth}
			\centering
			\includegraphics[width=2.8in]{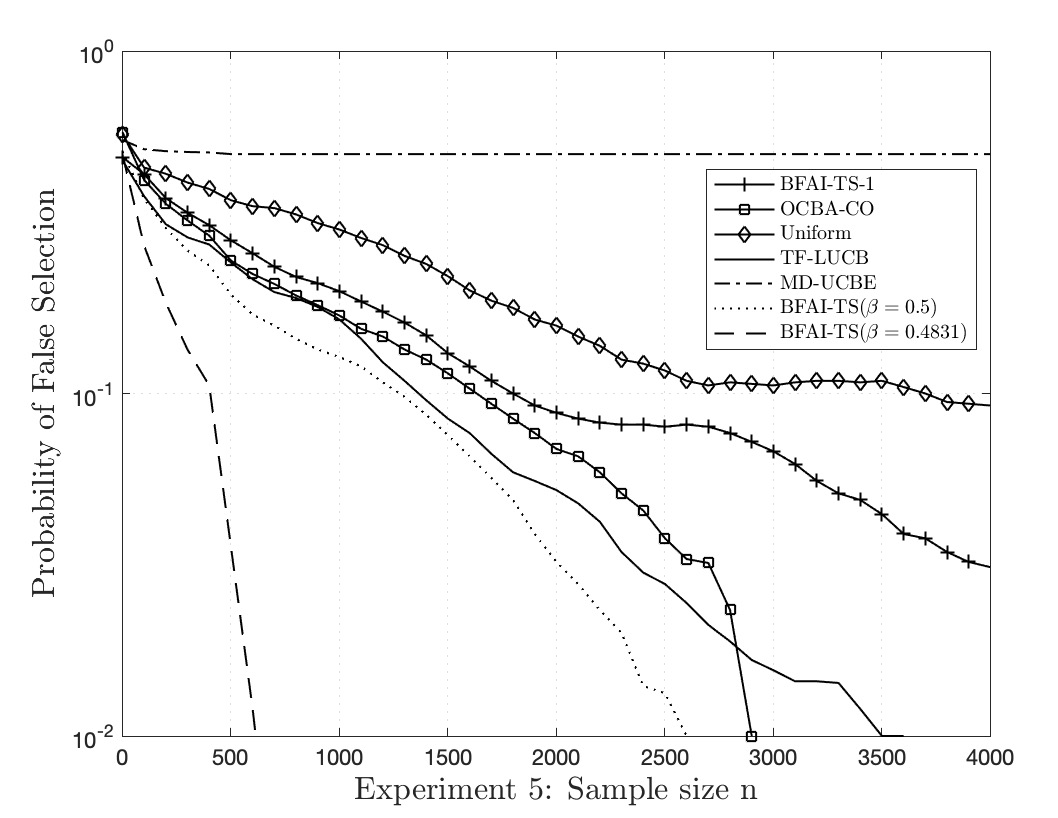}
		\end{minipage}%
	}%
	\subfigure[]{
		\begin{minipage}[t]{0.5\linewidth}
			\centering
			\includegraphics[width=2.8in]{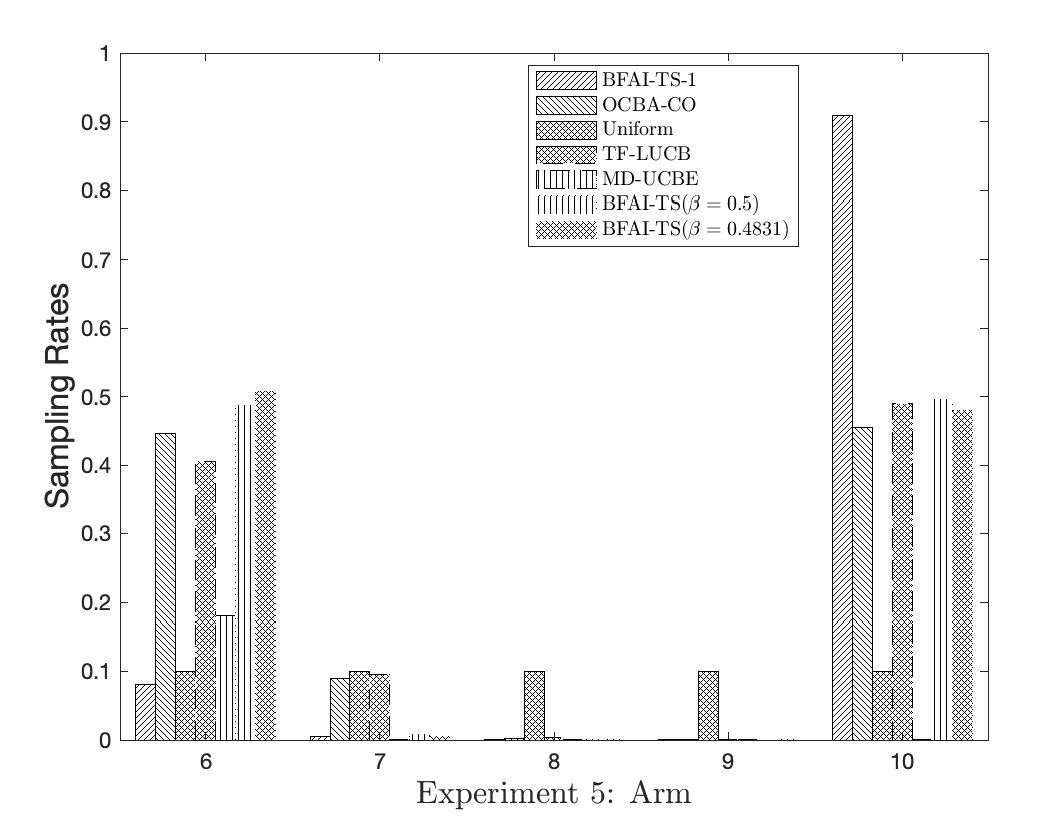}
		\end{minipage}%
	}%
	\centering
	\caption{PFS and their sampling rates on selected arms (Experiment 5)}
	
	\subfigure[]{
		\begin{minipage}[t]{0.5\linewidth}
			\centering
			\includegraphics[width=2.8in]{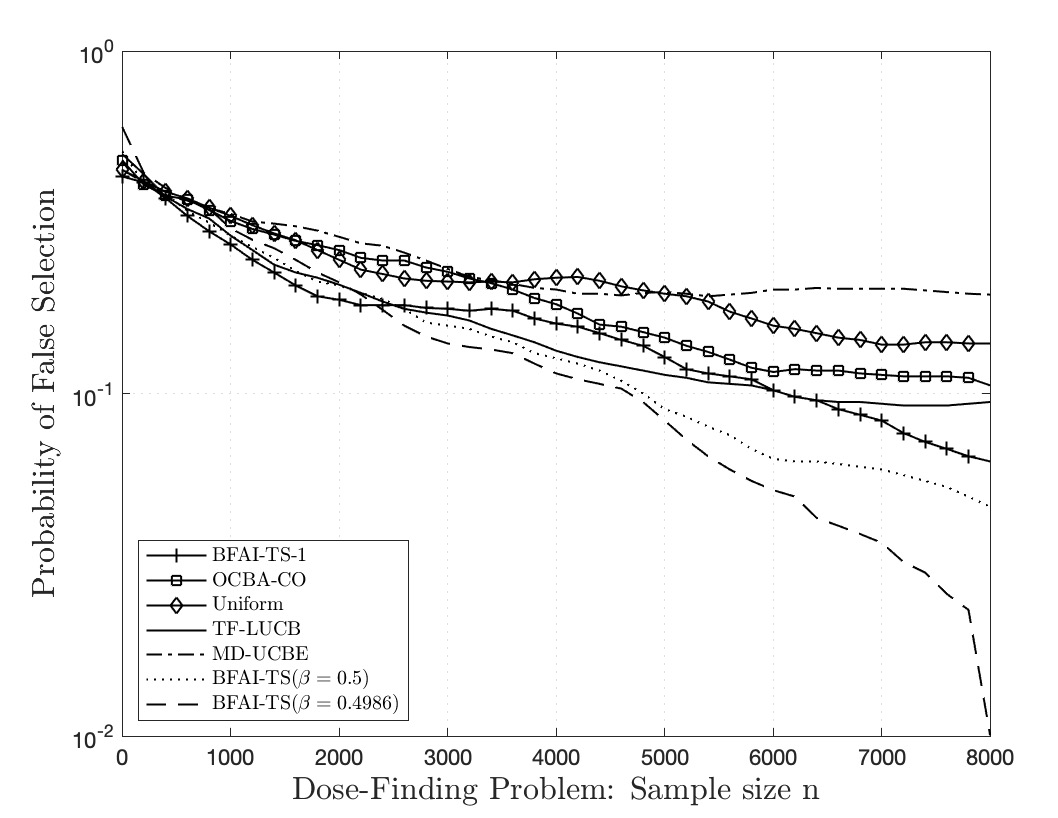}
		\end{minipage}%
	}%
	\subfigure[]{
		\begin{minipage}[t]{0.5\linewidth}
			\centering
			\includegraphics[width=2.8in]{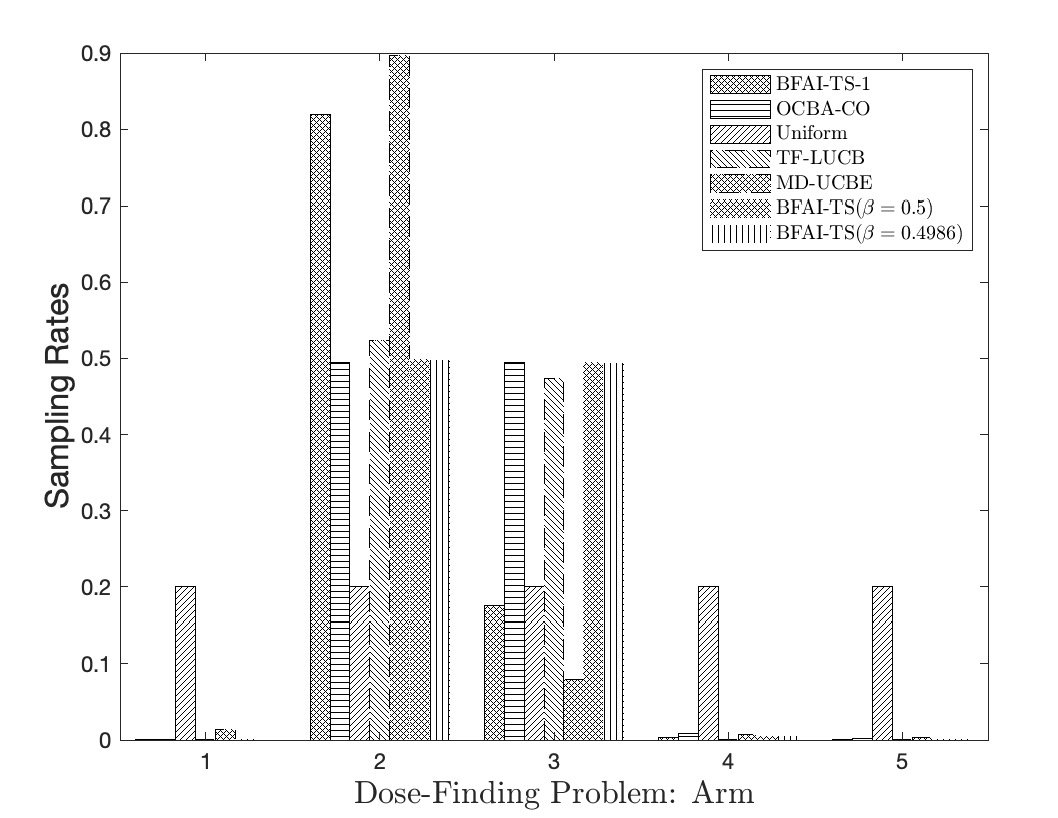}
		\end{minipage}%
	}%
	\centering
	\caption{PFS and their sampling rates on selected arms (Dose-finding problem)}
\end{figure*}

Figures 1(a)-6(a) show how probabilities of false selection (PFS) for the best feasible arms of the compared algorithms change with the sample sizes, and Figures 1(b)-6(b) show the sampling rates of the algorithms on selected arms. Specifically, we report sampling rates of arms $x=24, 25, 26, 27 \mbox{ and } 28$ for Experiments 1-4, arms $x=6, 7, 8, 9 \mbox{ and } 10$ for Experiments 5 and sampling rates of all five arms for the dose-finding problem. Figures 1(a)-6(a) align with the results reported in Table 1. When the values of $\beta$ are properly set, the proposed BFAI-TS performs the best, followed by OCBA-CO, TF-LUCB, BFAI-TS-1, Uniform and MD-UCBE. On the log scale, PFS of the proposed algorithm demonstrates a linear pattern, indicating the potential exponential rate of posterior convergence. When $\beta$ takes the optimal value in different experiments, the performance of BFAI-TS is slightly better than those when $\beta=0.5$. 

In Figures 1(b)-6(b), we can see that for the best feasible arm $x=26$ in Experiments 1-4, $x=10$ in Experiment 5 and $x=2$ in the dose-finding problem, the sampling rate is approximately equal to $\beta$. Compared to BFAI-TS, BFAI-TS-1 allocates too many samples to the best feasible arm. It indicates the necessity to adopt the top-two framework for the BFAI-TS algorithm. OCBA-CO allocates different proportions of samples with BFAI-TS due to the deviation when estimating the probability of the correct selection. MD-UCBE only investigates feasibility of the arms. It allocates too many samples to the arms for which feasibility detection is difficult. If a feasible arm is close to the constraint boundary and has poor objective performance, it will get more samples from MD-UCBE than from BFAI-TS. TF-LUCB only investigates arms that are not ruled out as infeasible arms. It allocates too many samples to the arms whose objective performances are better than that of the estimated best feasible arm, and arms whose infeasibility are difficult to identify.

\section{Conclusion}
In this research, we consider the problem of the best feasible arm identification (BFAI). BFAI extends BAI to the case with constraints. 
We focus on the fixed-budget setting and aim at minimizing the probability of falsely selecting the best feasible arm. We adopt the idea of Thompson sampling with adaptation to feasibility of the arm to solve BFAI. For the idea to be optimal, we further introduce a parameter to control the probabilities of sampling the best feasible arm and the set of remaining arms. The corresponding algorithm is called BFAI-TS. We establish asymptotic optimality of the algorithm in sample allocations and the rate of posterior convergence. The superior empirical performances of the proposed algorithm are demonstrated using numerical examples.
\bibliographystyle{plainnat} 
\bibliography{ref}

\appendix
\onecolumn
\section{Preliminaries}

\subsection{Notations}
\begin{itemize}
	\item[$k$] \phantom{======}total number of arms;
	\item[$n$] \phantom{======}total number of rounds;
	\item[$m$] \phantom{======}total number of constraint measures;
    \item[$X_{t,ij}$] \phantom{======}reward of arm $i$ and measure $j$ in round $t$, $i=1,2,\ldots,k$, $j=0,1,\ldots,m$, $t=$
    \item[\textcolor{white}{}] \phantom{======}$1,2,\ldots,n$;
	\item[$\mu_{ij}$] \phantom{======}true mean of arm $i$ and measure $j$, $i=1,2,\ldots,k$, $j=0,1,\ldots,m$;
    \item[$\sigma_{ij}^{2}$] \phantom{======}variance of arm $i$ and measure $j$, $i=1,2,\ldots,k$, $j=0,1,\ldots,m$;
    \item[$\gamma_{j}$] \phantom{======}threshold corresponding to measure $j$, $j=1,2,\ldots,m$;
    \item[$\mu_{t,ij}$] \phantom{======}posterior mean of arm $i$ and measure $j$ in round $t$, $i=1,2,\ldots,k$, $j=0,1,\ldots,m$, 
    \item[\textcolor{white}{}] \phantom{======}$t=1,2,\ldots,n$;
    \item[$\sigma_{t,ij}^{2}$] \phantom{======}posterior variance of arm $i$ and measure $j$ in round $t$, $i=1,2,\ldots,k$, $j=0,1,\ldots,m$, \item[\textcolor{white}{}] \phantom{======}$t=1,2,\ldots,n$;
    \item[$I_{t}$] \phantom{======}the arm which agent chooses to pull in round $t$, $t=1,2,\ldots,n$;
    \item[$N_{t,i}$] \phantom{======}number of samples for arm $i$ before round $t$, $i=1,2,\ldots,k$, $t=1,2,\ldots,n$;
    \item[$\theta_{ij}$] \phantom{======}variable of posterior mean of arm $i$ and measure $j$, $i=1,2,\ldots,k$, $j=0,1,\ldots,m$;
    \item[$P_{t,i}$] \phantom{======}the (posterior) probability that arm $i$ is the best feasible arm, $i=1,2,\ldots,k$, 
    \item[\textcolor{white}{}] \phantom{======}$t=1,2,\ldots,n$;
    \item[$\beta$] \phantom{======}parameter which controls the probability of sampling the best feasible arm; 
    \item[$I_{t}^{(1)}$] \phantom{======}the arm which the agent chooses with probability $\beta$ in round $t$, $t=1,2,\ldots,n$;
    \item[$I_{t}^{(2)}$] \phantom{======}the arm which the agent chooses with probability $1-\beta$ in round $t$, $t=1,2,\ldots,n$;
    \item[$\phi_{t,i}$] \phantom{======}the probability that the agent chooses arm $i$ in round $t$, $i=1,2,\ldots,k$, $t=1,$
    \item[\textcolor{white}{}] \phantom{======}$2,\ldots,n$;
    \item[$c_{t}$] \phantom{======}the probability that samples from all the arms are infeasible in round $t$, $t=1,$
    \item[\textcolor{white}{}] \phantom{======}$2,\ldots,n$;
    \item[$\mathcal{F}$] \phantom{======}the set of feasible arms;
    \item[$I_{n}^{*}$] \phantom{======}the estimated best feasible arm in round $n$;
    \item[$I^{*}$] \phantom{======}the best feasible arm;
    \item[$\mathcal{F}_{w}$] \phantom{======}the set of feasible but suboptimal arms;
    \item[$\mathcal{I}_{b}$] \phantom{======}the set of infeasible arms with objective performance no worse than $I^{*}$;
    \item[$\mathcal{I}_{w}$] \phantom{======}the set of infeasible arms with objective performance worse than $I^{*}$;
    \item[$\mathcal{M}_{F}^{i}$] \phantom{======}the set of constraints satisfied by arm $i$, $i=1,2\ldots,k$;
    \item[$\mathcal{M}_{I}^{i}$] \phantom{======}the set of constraints violated by arm $i$, $i=1,2\ldots,k$;
    \item[$\alpha_{i}^{\beta}$] \phantom{======}optimal sampling rates of arm $i$, when the proportion allocated to the best 
    \item[\textcolor{white}{}] \phantom{======}feasible arm is $\beta$, $i=1,2,\ldots,k$;
    \item[$\Gamma_{\beta}$] \phantom{======}optimal rate of posterior convergence when the proportion allocated to the best 
    \item[\textcolor{white}{}] \phantom{======}feasible arm is $\beta$;
    \item[$\Gamma_{\beta^{*}}$] \phantom{======}optimal rate of posterior convergence when the proportion allocated to the best \item[\textcolor{white}{}] \phantom{======}feasible arm $\beta$ is set as its optimal value $\beta^{*}$.
\end{itemize}
\subsection{Important Lemmas}
\begin{lemma}\label{lemma1}(\cite{qin2017improving} Lemma 1)
	Suppose random variable $X\sim \mathcal{N}(\mu,\sigma^{2})$ and constant $c>0$. Then
	\begin{equation*}
		\frac{1}{\sqrt{2\pi}}\exp\bigg(-\frac{(\sigma+c)^{2}}{2\sigma^{2}}\bigg)\leq \mathbb{P}(X\geq \mu+c)\leq \frac{1}{2} \exp\bigg(-\frac{c^{2}}{2\sigma^{2}}\bigg).
	\end{equation*}
\end{lemma}

\begin{lemma}\label{lemma2}(\cite{russo2020simple} Corollary 1)
	For $i\in A$, if $\sum_{t=2}^{n}\phi_{t,i}\rightarrow\infty$, then with probability $1$, $N_{n,i}\rightarrow\infty$ and
	\begin{equation*}
		\frac{\sum_{t=2}^{n}\phi_{t,i}}{N_{n,i}}\rightarrow1.
	\end{equation*} 
\end{lemma}

\section{Proof of Theorem 1}
In this section, we present the theoretical results specific to the sample allocations for the proposed BFAI-TS Algorithm. We first give a proposition showing that every arm will be sampled infinitely by the algorithm as the round $n$ goes to infinity.

\begin{proposition}\label{prop2}
	For the BFAI-TS Algorithm and sample path $w$, there exists a sample path dependent parameter $t(w)>0$ such that when $n>t(w)$, $I_{n}^{(1)}=I_{n}^{*}=1$.
\end{proposition}
Proposition \ref{prop2} indicates that when $n$ is large enough, the estimated best feasible arms of the algorithms become the real best feasible arms, i.e., $I_n^*=1$. In addition, the arms $I_n^{(1)}$ that the algorithms believe to be ``promising'' are also the real best feasible arms in the long run.

\subsection{Proof of Proposition 1}
To prove this proposition, we first define a set $V\triangleq\{i\in A: \sum_{t=2}^{\infty}\phi_{t,i}<\infty\}$, which contains arms receiving only a finite number of samples. By Lemma \ref{lemma2}, for all arms $i\in A$, $N_{\infty,i}<\infty$. Then for arm $i\notin V$ and measures $j= 0,1,2,\ldots m$, $|\mu_{n,ij}-\mu_{ij}|<\epsilon$ hold with probability $1$. It suffices to prove that $V=\emptyset$. With it, the claim in the proposition is straightforward by the Strong Law of Large Numbers. 

We will first show that $\lim_{n\rightarrow \infty}c_{n}\rightarrow0$. We prove this by contradiction. If $c_{n}>0$, then $\mathcal{F}\cap V^{c}=\emptyset$. Hence, by (1), $\phi_{n,i}>0$, which implies that $\sum_{t=2}^{\infty}\phi_{t,i}\rightarrow\infty$ for $i\in A$, i.e., $V=\emptyset$, which contradicts $c_{n}>0$ by the Strong Law of Large Numbers. Then the expression of $\phi_{t,i}$ simplifies as following
\begin{equation}
	\phi_{t,i}=P_{t,i}\beta+(1-\beta)P_{t,i}\sum_{i'\neq i}\Big(\frac{P_{t,i'}}{1-P_{t,i'}}\Big).
\end{equation}

Next we will show that if $V\neq \emptyset$, then $P_{n,i}>0$ for all arms $i\in V$. Let $\psi=\max_{i\in V^{c}\cap\mathcal{F}}\mu_{i0}$. For $i\in V$, we have
\begin{equation*}
\begin{split}
	&\mathbb{P}_{\theta\sim\Pi_{t}}\bigg((\theta_{i0}>\max_{\substack{i^{'}
    \neq i, \theta_{i^{'}j}\leq\gamma_{j},\forall j \in
    \{1,2,\ldots,m\}}}\theta_{i^{'}0}+\epsilon)\cap\bigcap_{j=1}^{m}(\theta_{ij}\leq\gamma_{j})\bigg)\\
	\geq&\mathbb{P}_{\theta\sim\Pi_{t}}\bigg((\theta_{i0}>\psi+2\epsilon)\cap \bigcap_{i'\in (V\cap \mathcal{F})\setminus\{i\}}\bigg((\theta_{i^{'}0}<\psi)\cap \bigcap_{j=1}^{m}\big((\theta_{i^{'}j}\leq\gamma_{j})\cap (\theta_{ij}\leq\gamma_{j})\big)\bigg)\bigg)\\
	&+\mathbb{P}_{\theta\sim\Pi_{t}}\bigg(\bigcap_{i'\in V\cap \mathcal{F}^{c}}\bigcup_{j=1}^{m}(\theta_{i^{'}j}>\gamma_{j})\bigg)-\mathbb{P}_{\theta\sim\Pi_{t}}\big(\max_{i\in V^{c}\cap\mathcal{F}}\theta_{i0}>\psi+\epsilon\big).\\
\end{split}
\end{equation*}
By the Strong Law of Large Numbers, $\mathbb{P}_{\theta\sim\Pi_{t}}\big(\max_{i\in V^{c}\cap\mathcal{F}}\theta_{i0}>\psi+\epsilon\big)\rightarrow0$ and $\mathbb{P}_{\theta\sim\Pi_{t}}\bigg((\theta_{i0}>\psi+2\epsilon)\cap \bigcap_{i'\in (V\cap \mathcal{F})\setminus\{i\}}\bigg((\theta_{i^{'}0}<\psi)\cap \bigcap_{j=1}^{m}\big((\theta_{i^{'}j}\leq\gamma_{j})\cap (\theta_{ij}\leq\gamma_{j})\big)\bigg)\bigg)+ \mathbb{P}_{\theta\sim\Pi_{t}}\bigg(\bigcap_{i'\in V\cap \mathcal{F}^{c}}\bigcup_{j=1}^{m}(\theta_{i^{'}j}>\gamma_{j})\bigg)>0$. Hence $P_{n,i}>0$ for all arms $i\in V$ and then $\sum_{t=2}^{\infty}\phi_{t,i}\rightarrow\infty$, which implies that $V=\emptyset$. Therefore, there exists a large enough $t(w)$ satisfying that when $n>t(w)$, $P_{n,1}\rightarrow 1$ and $P_{n,i}\rightarrow0$ for $i\neq 1$, i.e., $I_{n}^{(1)}=I_{n}^{*}=1$.

\subsection{Proof of Theorem 1}
We first prove the theorem for the best feasible arm $i=1$, i.e., there exists $T_{1}^{\epsilon}\triangleq\max\{t(w), \beta_{\max}t(w)/\epsilon\}$ such that for any $n\geq T_{1}^{\epsilon}$,	
\begin{equation*}
	\Bigg|\frac{\sum_{t=2}^{n}\phi_{t,1}}{n}-\beta\Bigg|\leq\epsilon,
\end{equation*}
where $\beta_{\max}=\max(\beta, 1-\beta)$.
Note that
\begin{equation*}
	\begin{split}
		\frac{\sum_{t=2}^{n}\phi_{t,1}}{n}=\frac{1}{n}\Bigg(\sum_{t=2}^{t(w)}\phi_{t,1}+\sum_{t=t(w)+1}^{n}\phi_{t,1}\Bigg)
		\leq\frac{1}{n}[\beta_{\max}(t(w)-1)+\beta(n-t(w))]
		<\beta+\frac{(\beta_{\max}-\beta)t(w)}{n},
	\end{split}
\end{equation*}
and 
\begin{equation*}
	\begin{split}
		\frac{\sum_{t=2}^{n}\phi_{t,1}}{n}=\frac{1}{n}\Bigg(\sum_{t=2}^{t(w)}\phi_{t,1}+\sum_{t=t(w)+1}^{n}\phi_{t,1}\Bigg)
		\geq\frac{1}{n}\beta(n-t(w))
		=\beta-\frac{\beta t(w)}{n}.
	\end{split}
\end{equation*}
For any $n\geq T_{1}^{\epsilon}$, we can obtain $|\frac{\sum_{t=2}^{n}\phi_{t,1}}{n}-\beta|\leq\epsilon$.

We next prove the theorem for the non-best-feasible arms $i=2,3,\ldots,k$, i.e., with parameter $\beta\in(0,1)$, there exists $T_{2}^{\epsilon}=\max(t(w), 1/\epsilon, \epsilon)$ such that for any $n\geq T_{2}^{\epsilon}$
\begin{equation*}
	\Bigg|\frac{N_{n,i}}{n}-\alpha_{i}^{\beta}\Bigg|\leq\epsilon\quad\forall i\neq 1,
\end{equation*}
where the unique vector $(\alpha_{2}^{\beta}, \ldots, \alpha_{k}^{\beta})$ satisfies 
\begin{equation}
	\sum_{i=2}^{k}\alpha_{i}^{\beta}=1-\beta, \text{ and } \mathcal{R}_i=\mathcal{R}_{i'} \text{ for any }i\neq i^{'}\neq 1,
\end{equation}
where $\mathcal{R}_i=\frac{(\mu_{i0}-\mu_{10})^{2}}{(\sigma_{i0}^{2}/\alpha_{i}^{\beta}+\sigma_{10}^{2}/\beta)}\mathbf{1}\{i\in\mathcal{F}_{w}\cup \mathcal{I}_{w}\}	+\alpha_{i}^{\beta}\sum\limits_{j\in\mathcal{M}_{I}^{i}}\frac{(\mu_{ij}-\gamma_{j})^{2}}{\sigma_{ij}^{2}}\mathbf{1}\{i\in\mathcal{I}_{b}\cup \mathcal{I}_{w}\}$. 

For $\phi_{n,i}$, we have
\begin{equation*}
	\phi_{n,i}\leq P_{n,i}\beta+(1-\beta)P_{n,i}\frac{\sum_{i'\neq i}P_{n,i'}}{1-P_{n,1}}\leq  P_{n,i}\beta+(1-\beta)\frac{P_{n,i}}{\max_{i\neq 1}P_{n,i}}\leq \frac{P_{n,i}}{\max_{i\neq 1}P_{n,i}}.
\end{equation*}
Note that $\mathbb{P}\{FE_{i}\}=P_{n,i}$ for $i\neq 1$. Suppose there is an arm $i$ whose sampling ratio satisfies $N_{n,i}/n-\alpha_{i}^{\beta}>\epsilon$ and an arm $i'$ whose sampling ratio satisfies $N_{n,i'}/n<\alpha_{i'}^{\beta}$. Then
\begin{equation*}
	\begin{split}
		&\phi_{n,i}\leq \frac{P_{n,i}}{\max_{i\neq 1}P_{n,i}}\leq \frac{P_{n,i}}{P_{n,i'}}\\
		\dot{=}&\frac{\exp(-n({\frac{(\mu_{i0}-\mu_{10}-2\epsilon)^{2}}{(\sigma_{i0}^{2}/(\alpha_{i}^{\beta}+\epsilon)+\sigma_{10}^{2}/\beta)}\mathbf{1}\{i\in\mathcal{F}_{w}\cup \mathcal{I}_{w}\}
				+(\alpha_{i}^{\beta}+\epsilon)\sum\limits_{j\in\mathcal{M}_{I}^{i}}\frac{(\mu_{ij}-\gamma_{j}-\epsilon)^{2}}{\sigma_{ij}^{2}}\mathbf{1}\{i\in\mathcal{I}_{b}\cup \mathcal{I}_{w}\}}))}{\exp(-n(\frac{(\mu_{i'0}-\mu_{10}+2\epsilon)^{2}}{(\sigma_{i'0}^{2}/\alpha_{i'}^{\beta}+\sigma_{10}^{2}/\beta)}\mathbf{1}\{i'\in\mathcal{F}_{w}\cup \mathcal{I}_{w}\}
			+\alpha_{i'}^{\beta}\sum\limits_{j\in\mathcal{M}_{I}^{i'}}\frac{(\mu_{i'j}-\gamma_{j}+\epsilon)^{2}}{\sigma_{i'j}^{2}}\mathbf{1}\{i'\in\mathcal{I}_{b}\cup \mathcal{I}_{w}\}))}.
	\end{split}
\end{equation*}
Notice that molecules gradually decreases to $\mathcal{R}_{n,i}$ as $\epsilon$ decreases to $0$ and denominator  gradually increases to $\mathcal{R}_{n,i'}$ as $\epsilon$ decreases to $0$. Then we can find $\delta>0$ such that $\frac{(\mu_{i0}-\mu_{10}+2\epsilon)^{2}}{(\sigma_{i0}^{2}/(\alpha_{i}^{\beta}+\epsilon)+\sigma_{10}^{2}/\beta)}\mathbf{1}\{i\in\mathcal{F}_{w}\cup \mathcal{I}_{w}\}
+(\alpha_{i}^{\beta}+\epsilon)\sum\limits_{j\in\mathcal{M}_{I}^{i}}\frac{(\mu_{ij}-\gamma_{j}-\epsilon)^{2}}{\sigma_{ij}^{2}}$
$\cdot\mathbf{1}\{i\in\mathcal{I}_{b}\cup \mathcal{I}_{w}\}>\mathcal{R}_{n,i}+\delta$. Hence
\begin{equation*}
	\phi_{n,i}\leq \frac{\exp(-n(\mathcal{R}_{n,i}+\delta))}{\exp(-n\mathcal{R}_{n,i'})}\leq \exp(-n\delta).
\end{equation*}
It implies that if arm $i$ is pulled for too many rounds, the probability that it is pulled in the next round is negligible. Then
\begin{equation*}
	\sum_{t=t(w)}^{\infty}\phi_{t,i}\mathbf{1}\{\frac{N_{n,i}}{n}>\alpha_{i}^{\beta}+\epsilon\}\leq \sum_{t=t(w)}^{\infty} \exp(-n\delta)<\infty.
\end{equation*}
We have
\begin{equation*}
	\sum_{t=2}^{\infty}\phi_{t,i}=\sum_{t=2}^{t(w)}\phi_{t,i}+\sum_{t=t(w)+1}^{\infty}\phi_{t,i}\mathbf{1}\{\frac{N_{n,i}}{n}>\alpha_{i}^{\beta}+\epsilon\}+\sum_{t=t(w)+1}^{\infty}\phi_{t,i}\mathbf{1}\{\frac{N_{n,i}}{n}<\alpha_{i}^{\beta}+\epsilon\}\rightarrow\infty.
\end{equation*}
Hence $N_{n,i}/n<\alpha_{i}^{\beta}+\epsilon$. Suppose there is an arm $i$ satisfying $N_{n,i}/n<\alpha_{i}^{\beta}-\epsilon$ and for arm $i'\neq i$, $N_{n,i'}/n<\alpha_{i'}^{\beta}+\epsilon/k$. On the one hand, $\sum_{i'\in A}N_{n,i'}/n=\sum _{i'\neq i }N_{n,i'}/n+N_{n,i}/n<\sum_{i'\neq i}\alpha_{i'}^{\beta}+\epsilon/k+\alpha_{i}^{\beta}-\epsilon=1-\epsilon/k$. On the other hand, $\sum_{i'\in A}N_{n,i'}/n=n-1/n>1-\epsilon/k$, which leads to $|N_{n,i}-\alpha_{i}^{\beta}|<\epsilon$.

\section{Proof of Theorem 2}
In this section, we present the theoretical results specific to the asymptotic optimality in the rate of posterior convergence for the proposed BFAI-TS Algorithm. During the proof process, values or sets indicated by superscripts or subscripts with $t$ or $n$ represent the estimates of those values or sets in the round $t$ or $n$, respectively.

After $n$ rounds, the probability that the best feasible arm (arm 1) is correctly identified is given by
\begin{equation}\label{(1)}
	\begin{split}
		P_{t,1}= \mathbb{P}_{\theta\sim\Pi_{t}}\Bigg(\bigcap_{i'\neq 1}\Big((\theta_{10}<\theta_{i^{'}0})\cap
        \bigcap_{j=1}^{m}(\theta_{i^{'}j}\leq\gamma_{j})\Big)^{c}
        \cap\bigcap_{j=1}^{m}(\theta_{1j}\leq\gamma_{j})\Bigg),
	\end{split}
\end{equation}
which is the probability that arm 1 is estimated feasible and objectively superior. Then, $1-P_{n,1}$ is the probability of false selection after $n$ rounds with $1-P_{n,1}=\mathbb{P}_{\theta\sim \Pi_{n}}\Big(FE_{1}\cup\bigcup\limits_{i\neq1}FE_{i}\Big)$, where $FE_1=\bigcup\limits_{j=1}^{m}(\theta_{1j}>\gamma_{j})$ represents the event of false evaluation of arm 1, i.e., arm 1 is estimated infeasible, and $FE_i=(\theta_{i0}\geq \theta_{10})\cap\bigcap\limits_{j=1}^{m}(\theta_{ij}\leq\gamma_{j})\cap\bigcap\limits_{j=1}^{m}(\theta_{1j}\leq\gamma_{j})$ for $i\neq 1$ represents the event of false evaluation of arm $i$, i.e., arm 1 is estimated feasible and arm $i$ is estimated feasible and objectively superior. Note that $\mathbb{P}\{FE_{i}\}= P_{n,i}$ for $i\neq 1$.

For real-valued sequences $\{a_n\}$ and $\{b_n\}$, we call them logarithmically equivalent if $\lim_{n\rightarrow\infty}\frac{1}{n}\log\frac{a_{n}}{b_{n}}=0$ and denote it by $a_{n}\dot{=}b_{n}$. Since $\max(\mathbb{P}\{FE_{1}\},\max_{i\neq1}\mathbb{P}\{FE_{i}\})\leq1-P_{n,1}\leq k\max(\mathbb{P}\{FE_{1}\},\max_{i\neq1}\mathbb{P}\{FE_{i}\})$,
we have
\begin{equation}\label{5.4}
	\begin{split}
		1-P_{n,1}\dot{=}&\max(\mathbb{P}\{FE_{1}\},\max_{i\neq1}\mathbb{P}\{FE_{i}\})
		\dot{=}\max_{i\in\{1,2,...,k\}}\mathbb{P}\{FE_i\}.
	\end{split}
\end{equation}
To this end, $FE_i$ for $i=1,2,...,k$ can be treated as the contribution from arm $i$ to the false selection event, and the rate of posterior convergence for the probability of false selection is governed by the largest probability of false evaluation of the $k$ arms. Therefore, to develop an algorithm for BFAI, it makes sense to iteratively pull the arm with the largest probability of false evaluation $\mathbb{P}\{FE_i\}$, i.e., the arm that contributes the most to the probability of false selection.

Note that $\mathbb{P}\{FE_i\}$ does not have an analytical form. Next, we discuss how to approximately calculate $\mathbb{P}\{FE_i\}$, $\max_{i\in\{1,2,...,k\}} \mathbb{P}\{FE_i\}$ and $1-P_{n,1}$ in (\ref{5.4}). We introduce the following lemma.

\begin{lemma}\label{newadd1}
	The probabilities of false evaluation $\mathbb{P}\{FE_{1}\}$ of arm 1 and $\mathbb{P}\{FE_{i}\}$ of arm $i$ for $i \neq 1$ are logarithmically equivalent to
	\begin{align*}
		\mathbb{P}\{FE_{1}\}\dot{=}&\max\limits_{j\in\mathcal{M}_{F}^{1,n}}\exp\bigg({-\frac{(\gamma_{j}-\mu_{n,1j})^{2}}{2\sigma_{1j}^{2}/N_{n,1}}}\bigg),\\
		\mathbb{P}\{FE_{i}\}
		\dot{=}&\exp\bigg(-\frac{(\mu_{n,i0}-\mu_{n,10})^{2}}{2(\sigma_{i0}^{2}/N_{n,i}+\sigma_{10}^{2}/N_{n,1})}\mathbf{1}\{i\in\mathcal{F}_{w}^{n}\cup \mathcal{I}_{w}^{n}\}\bigg) 
        \exp\bigg({-\sum\limits_{j\in\mathcal{M}_{I}^{i,n}}\frac{(\gamma_{j}-\mu_{n,ij})^{2}}{2\sigma_{ij}^{2}/N_{n,i}}}\mathbf{1}\{i\in\mathcal{I}_{b}^{n}\cup \mathcal{I}_{w}^{n}\}\bigg).
	\end{align*}
	Note that arm $i\neq 1$ must fall in the set $\mathcal{F}_{w}^{n}\cup \mathcal{I}_{w}^{n}$ or $\mathcal{I}_{b}^{n}\cup \mathcal{I}_{w}^{n}$. As a result,
	\begin{align*}
		1-P_{n,1}
		\dot{=}&\exp\Bigg(-n\min\limits_{i\in A}\bigg(\frac{(\mu_{n,i0}-\mu_{n,10})^{2}}{2(\sigma_{i0}^{2}n/N_{n,i}+\sigma_{10}^{2}n/N_{n,1})}
         \mathbf{1}\{i\in\mathcal{F}^{n}\cup \mathcal{I}_{w}^{n}\} +\sum\limits_{j\in\mathcal{M}_{I}^{i,n}}\frac{(\gamma_{j}-\mu_{n,ij})^{2}}{2\sigma_{ij}^{2}n/N_{n,i}}
         \mathbf{1}\{i\in \mathcal{I}_{b}^{n}\cup \mathcal{I}_{w}^{n}\}
		\bigg)\Bigg).
	\end{align*}
\end{lemma}

Lemma \ref{newadd1} indicates that the probability of false evaluation of arm 1 can be represented by the probability that arm 1 is falsely identified as violating the constraint that it is most likely to violate, and the probability of false evaluation of arm $i$ for $i\neq 1$ can be represented by the probability that this arm is falsely identified as feasible or the probability that this arm is falsely identified as objectively superior, whichever is larger. 

\subsection{Proof of Lemma \ref{newadd1}}
We first analyze $\mathbb{P}\{FE_1\}$. Note that in round $n$, $\theta_{1j}-\gamma_{j}\sim\mathcal{N}(\mu_{n,1j}-\gamma_{j},\sigma_{1j}^{2}/N_{n,1})$. Since arm 1 is feasible, when $n$ is large enough, $\mu_{n,1j}\leq\gamma_{j}$. Then 
\begin{equation*}
	\max\limits_{j\in\mathcal{M}_{F}^{1,n}}\mathbb{P}_{\theta\sim\Pi_{n}}(\theta_{1j}>\gamma_{j})\leq \mathbb{P}\{FE_{1}\}=\mathbb{P}_{\theta\sim\Pi_{n}}\bigg(\bigcup\limits_{j=1}^{m}(\theta_{1j}>\gamma_{j})\bigg)\leq m\max\limits_{j\in\mathcal{M}_{F}^{1,n}}\mathbb{P}_{\theta\sim\Pi_{n}}(\theta_{1j}>\gamma_{j}),
\end{equation*}
so
\begin{equation*}
	\mathbb{P}\{FE_{1}\}\dot{=}\max\limits_{j\in\mathcal{M}_{F}^{1,n}}\mathbb{P}(\theta_{1j}>\gamma_{j}),
\end{equation*}
where $\dot{=}$ (logarithmically equivalent) has been defined in Section 4 of the paper. By Lemma \ref{lemma1}, we have
\begin{equation*}
	\frac{1}{\sqrt{2\pi}}\exp\Bigg(-\frac{(\sigma_{1j}/{\sqrt{N_{n,1}}}+\gamma_{j}-\mu_{n,1j})^{2}}{2\sigma_{1j}^{2}/N_{n,1}}\Bigg)\leq \mathbb{P}_{\theta\sim\Pi_{n}}(\theta_{1j}-\gamma_{j}>0)\leq\frac{1}{2}\exp\Bigg(-\frac{(\gamma_{j}-\mu_{n,1j})^{2}}{2\sigma_{1j}^{2}/N_{n,1}}\Bigg),
\end{equation*}
which implies 
\begin{equation*}
	\frac{1}{n}\log\Big(\frac{1}{\sqrt{2\pi}}\Big)-\frac{1}{2n}-\frac{\gamma_{j}-\mu_{n,1j}}{n\sqrt{\sigma_{1j}^{2}/N_{n,1}}}\leq
	\frac{1}{n}\log\Bigg(\frac{\mathbb{P}_{\theta\sim\Pi_{n}}(\theta_{1j}-\gamma_{j}>0)}{\exp\Big({-\frac{(\gamma_{j}-\mu_{n,1j})^{2}}{2\sigma_{1j}^{2}/N_{n,1}}}\Big)}\Bigg)
	\leq\frac{1}{n}\log\Big(\frac{1}{2}\Big).
\end{equation*}
Note that when $\mu_{n,1j}\leq\gamma_{j}$,
\begin{equation*}
	0\leq\frac{\gamma_{j}-\mu_{n,1j}}{n\sqrt{\sigma_{1j}^{2}/N_{n,1}}}=\frac{\gamma_{j}-\mu_{n,1j}}{\sigma_{1j}\sqrt{n\frac{n}{N_{n,1}}}}\leq\frac{\gamma_{j}-\mu_{n,1j}}{\sigma_{1j}\sqrt{n}}.
\end{equation*}
Using the Squeeze Theorem, we have
\begin{equation*}
	\lim\limits_{n\rightarrow\infty}\frac{\gamma_{j}-\mu_{n,1j}}{n\sqrt{\sigma_{1j}^{2}/N_{n,1}}}=0
\end{equation*}
and
\begin{equation*}
	\lim\limits_{n\rightarrow\infty}	\frac{1}{n}\log\Bigg(\frac{\mathbb{P}_{\theta\sim\Pi_{n}}(\theta_{1j}-\gamma_{j}>0)}{\exp\Big({-\frac{(\gamma_{j}-\mu_{n,1j})^{2}}{2\sigma_{1j}^{2}/N_{n,1}}}\Big)}\Bigg)=0.
\end{equation*}
Then we can obtain
\begin{equation*}
	\mathbb{P}_{\theta\sim\Pi_{n}}(\theta_{1j}>\gamma_{j})\dot{=}\exp\Bigg({-\frac{(\gamma_{j}-\mu_{n,1j})^{2}}{2\sigma_{1j}^{2}/N_{n,1}}}\Bigg).
\end{equation*}
Hence 
\begin{equation}
	\mathbb{P}\{FE_{1}\}\dot{=}\max\limits_{j\in\mathcal{M}_{F}^{1,n}}\exp\Bigg({-\frac{(\gamma_{j}-\mu_{n,1j})^{2}}{2\sigma_{1j}^{2}/N_{n,1}}}\Bigg).
\end{equation}

Next, we analyze $\mathbb{P}\{FE_{i}\}$ for $i=2,3,\ldots,k$.
\begin{equation*}
	\mathbb{P}\{FE_{i}\}=\mathbb{P}_{\theta\sim\Pi_{n}}\Bigg((\theta_{i0}\geq \theta_{10})\cap\bigcap\limits_{j=1}^{m}(\theta_{ij}\leq\gamma_{j})\cap\bigcap\limits_{j=1}^{m}(\theta_{1j}\leq\gamma_{j})\Bigg).
\end{equation*}
Since the probability that arm 1 is estimated feasible and the probability that arm $i$ is estimated feasible on constraints $j\in\mathcal{M}_{F}^{i,n}$ both tend to $1$ as $n$ goes to infinity, we have
\begin{equation}
	\mathbb{P}\{FE_{i}\}=\mathbb{P}_{\theta\sim\Pi_{n}}\big(\theta_{i0}\geq \theta_{10})\quad \mbox{for~}i\in \mathcal{F}_{w}^{n},
\end{equation}
\begin{equation}
	\mathbb{P}\{FE_{i}\}=\mathbb{P}_{\theta\sim\Pi_{n}}\Big(\bigcap\limits_{j\in\mathcal{M}_{I}^{i,n}}(\theta_{ij}\leq\gamma_{j})\Big)\quad\mbox{for~}i\in \mathcal{I}_{b}^{n},
\end{equation}
\begin{equation}
	\mathbb{P}\{FE_{i}\}=\mathbb{P}_{\theta\sim\Pi_{n}}\Big((\theta_{i0}\geq \theta_{10})\cap\bigcap\limits_{j\in\mathcal{M}_{I}^{i,n}}(\theta_{ij}\leq\gamma_{j})\Big) \quad\mbox{for~}i\in \mathcal{I}_{w}^{n}.
\end{equation}
Note that in round $n$, $\theta_{ij}-\gamma_{j}\sim\mathcal{N}(\mu_{n,ij}-\gamma_{j},\sigma_{ij}^{2}/N_{n,i})$ and $\theta_{i0}-\theta_{10}\sim\mathcal{N}(\mu_{n,i0}-\mu_{n,10},\sigma_{i0}^{2}/{N_{n,i}}+\sigma_{10}^{2}/{N_{n,1}})$. Since arm $i$ violates constraint $j$ in set $\mathcal{M}_{I}^{i,n}$, when $n$ is large enough, $\mu_{n,ij}>\gamma_{j}$. Also, since arm $i$ in set $\mathcal{F}_{w}^{n}$ has worse objective performance than arm 1, when $n$ is large enough, $\mu_{n,10}\geq \mu_{n,i0}$. Then 
\begin{equation*}
	\frac{1}{\sqrt{2\pi}}\exp\Bigg(-\frac{(\sigma_{ij}/{\sqrt{N_{n,i}}}+\mu_{n,ij}-\gamma_{j})^{2}}{2\sigma_{ij}^{2}/N_{n,i}}\Bigg)\leq \mathbb{P}_{\theta\sim\Pi_{n}}(\gamma_{j}\geq\theta_{ij})\leq\frac{1}{2}\exp\Bigg(-\frac{(\mu_{n,ij}-\gamma_{j})^{2}}{2\sigma_{ij}^{2}/N_{n,i}}\Bigg),
\end{equation*}
and
\begin{equation*}
	\begin{split}
		\frac{1}{\sqrt{2\pi}}\exp\Bigg(-\frac{(\sqrt{\sigma_{i0}^{2}/{N_{n,i}}+\sigma_{10}^{2}/{N_{n,1}}}+\mu_{n,10}-\mu_{n,i0})^{2}}{2(\sigma_{i0}^{2}/{N_{n,i}}+\sigma_{10}^{2}/{N_{n,1}})}\Bigg)\leq \mathbb{P}_{\theta\sim\Pi_{n}}(\theta_{i0}>\theta_{10})
		\leq\frac{1}{2}\exp\Bigg(-\frac{(\mu_{n,10}-\mu_{n,i0})^{2}}{2(\sigma_{i0}^{2}/{N_{n,i}}+\sigma_{10}^{2}/{N_{n,1}})}\Bigg).
	\end{split}
\end{equation*}
Similarly, we know
\begin{equation*}
	\mathbb{P}_{\theta\sim\Pi_{n}}\Big(\bigcap\limits_{j\in\mathcal{M}_{I}^{i,n}}(\theta_{ij}\leq\gamma_{j})\Big)=\prod\limits_{j\in\mathcal{M}_{I}^{i,n}}\mathbb{P}_{\theta\sim\Pi_{n}}(\theta_{ij}\leq\gamma_{j})\dot{=}\exp\Bigg({-\sum\limits_{j\in\mathcal{M}_{I}^{i,n}}\frac{(\gamma_{j}-\mu_{n,ij})^{2}}{2\sigma_{ij}^{2}/N_{n,i}}}\Bigg),
\end{equation*}
and
\begin{equation*}
	\mathbb{P}_{\theta\sim\Pi_{n}}(\theta_{i0}\geq \theta_{10})\dot{=}\exp\Bigg(-\frac{(\mu_{n,i0}-\mu_{n,10})^{2}}{2(\sigma_{i0}^{2}/N_{n,i}+\sigma_{10}^{2}/N_{n,1})}\Bigg).
\end{equation*}
Due to independence of samples across different measures,
\begin{equation}
	\begin{split}
		\mathbb{P}\{FE_{i}\}\dot{=}&\exp\Bigg({-\sum\limits_{j\in\mathcal{M}_{I}^{i,n}}\frac{(\gamma_{j}-\mu_{n,ij})^{2}}{2\sigma_{ij}^{2}/N_{n,i}}}\mathbf{1}\{i\in\mathcal{I}_{b}^{n}\cup \mathcal{I}_{w}^{n}\}\Bigg)
		\exp\Bigg(-\frac{(\mu_{n,i0}-\mu_{n,10})^{2}}{2(\sigma_{i0}^{2}/N_{n,i}+\sigma_{10}^{2}/N_{n,1})}\mathbf{1}\{i\in\mathcal{F}_{w}^{n}\cup \mathcal{I}_{w}^{n}\}\Bigg).
	\end{split}
\end{equation}
Next, we analyze $\max_{i\in\{1,2,\ldots,k\}} \mathbb{P}\{FE_i\}$.
\begin{equation}\label{5.9}
	\begin{split}
		&1-P_{n,1}\\
        \dot{=}&\max_{i\in\{1,2,…,k\}}\mathbb{P}\{FE_i\}\\
		\dot{=}&\max\Bigg(\max\limits_{i\neq1}\exp\bigg(-\frac{(\mu_{n,i0}-\mu_{n,10})^{2}}{2(\sigma_{i0}^{2}/N_{n,i}+\sigma_{10}^{2}/N_{n,1})}\mathbf{1}\{i\in \mathcal{F}_{w}^{n}\cup \mathcal{I}_{w}^{n}\}\bigg)
		\exp\bigg({-\sum\limits_{j\in\mathcal{M}_{I}^{i,n}}\frac{(\gamma_{j}-\mu_{n,ij})^{2}}{2\sigma_{ij}^{2}/N_{n,i}}}\mathbf{1}\{i\in \mathcal{I}_{b}^{n}\cup \mathcal{I}_{w}^{n}\}\bigg),\\
		&\max\limits_{j\in\mathcal{M}_{F}^{1,n}}\exp\bigg({-\frac{(\gamma_{j}-\mu_{n,1j})^{2}}{2\sigma_{1j}^{2}/N_{n,1}}}\bigg)\Bigg)\\
		\dot{=}&\exp\Bigg(-n\min\bigg(\min\limits_{i\neq1}\Big(\frac{(\mu_{n,i0}-\mu_{n,10})^{2}}{2(\sigma_{i0}^{2}n/N_{n,i}+\sigma_{10}^{2}n/N_{n,1})}\mathbf{1}\{i\in \mathcal{F}_{w}^{n}\cup \mathcal{I}_{w}^{n}\}
		+\sum\limits_{j\in\mathcal{M}_{I}^{i,n}}\frac{(\gamma_{j}-\mu_{n,ij})^{2}}{2\sigma_{ij}^{2}n/N_{n,i}}
		\mathbf{1}\{i\in \mathcal{I}_{b}^{n}\cup \mathcal{I}_{w}^{n}\}\Big),\\ &\min\limits_{j\in\mathcal{M}_{F}^{1,n}}\frac{(\gamma_{j}-\mu_{n,1j})^{2}}{2\sigma_{1j}^{2}n/N_{n,1}}\bigg)\Bigg).
	\end{split}
\end{equation}
To further analyze (\ref{5.9}), let $\mathbf{\Omega}\triangleq\{\mathbf{\alpha}=(\alpha_{1},\alpha_{2}\ldots,\alpha_{k}):\sum_{i=1}^{k}\alpha_{i}=1 \mbox{~and~} \alpha_{i}\geq0, \forall i\in A\}$ denote the set of feasible sampling rates for the $k$ arms. 

We investigate the rate function
\begin{equation}
	\begin{split}
		&\lim\limits_{n\rightarrow\infty}-\frac{1}{n}\log(1-P_{n,1})\\
		\dot{=}&\lim\limits_{n\rightarrow\infty}\min\Bigg(\min_{i\neq 1}\Big(\frac{(\mu_{n,i0}-\mu_{n,10})^{2}}{2(\sigma_{i0}^{2}n/N_{n,i}+\sigma_{10}^{2}n/N_{n,1})}\mathbf{1}\{i\in \mathcal{F}_{w}^{n}\cup \mathcal{I}_{w}^{n}\}
		+\sum\limits_{j\in\mathcal{M}_{I}^{i,n}}\frac{(\gamma_{j}-\mu_{n,ij})^{2}}{2\sigma_{ij}^{2}n/N_{n,i}}\mathbf{1}\{i\in \mathcal{I}_{b}^{n}\cup \mathcal{I}_{w}^{n}\}\Big),\\
        &\min\limits_{j\in\mathcal{M}_{F}^{1,n}}\frac{(\gamma_{j}-\mu_{n,1j})^{2}}{2\sigma_{1j}^{2}n/N_{n,1}}\Bigg)\\
		=&\min\bigg(
		\min_{i\neq 1}\Big(
		\frac{(\mu_{i0}-\mu_{10})^{2}}{2(\sigma_{i0}^{2}/\alpha_{i}+\sigma_{10}^{2}/\alpha_{1})}\mathbf{1}\{i\in \mathcal{F}_{w}\cup \mathcal{I}_{w}\}
		+\alpha_{i}\sum\limits_{j\in\mathcal{M}_{I}^{i}}\frac{(\mu_{ij}-\gamma_{j})^{2}}{2\sigma_{ij}^{2}}\mathbf{1}\{i\in \mathcal{I}_{b}\cup \mathcal{I}_{w}\}\Big),\\
		&\min\limits_{j\in\mathcal{M}_{F}^{1}}\alpha_{1}\frac{(\mu_{1j}-\gamma_{j})^{2}}{2\sigma_{1j}^{2}}\bigg).
	\end{split}
\end{equation}
If $\lim\limits_{n\rightarrow\infty}-\frac{1}{n}\log(1-P_{n,1})\dot{=} \min\limits_{j\in\mathcal{M}_{F}^{1}}\alpha_{1}\frac{(\mu_{1j}-\gamma_{j})^{2}}{2\sigma_{1j}^{2}}$, it is a monotonically increasing function of $\alpha_{1}$, so we can improve the rate of posterior convergence for false evaluation of arm $1$ by increasing its sampling rate $\alpha_1$. Note that if we keep increasing $\alpha_1$, $\lim\limits_{n\rightarrow\infty}-\frac{1}{n}\log(1-P_{n,1})\dot{=} \min\limits_{j\in\mathcal{M}_{F}^{1}}\alpha_{1}\frac{(\mu_{1j}-\gamma_{j})^{2}}{2\sigma_{1j}^{2}}$ will increase, and there will be a point such that $\lim\limits_{n\rightarrow\infty}-\frac{1}{n}\log(1-P_{n,1})\dot{=} \min\limits_{j\in\mathcal{M}_{F}^{1}}\alpha_{1}\frac{(\mu_{1j}-\gamma_{j})^{2}}{2\sigma_{1j}^{2}}$ does not hold any more. Let $\tilde{\alpha}_{1}$ be the maximum value of $\alpha_1$ such that the equation still holds. Then,
\begin{equation*}
	\begin{split}
		&\max_{\alpha_{1}\in(0,\tilde{\alpha}_{1}]}\lim\limits_{n\rightarrow\infty}-\frac{1}{n}\log(1-P_{n,1}) \leq \tilde{\alpha}_{1} \min\limits_{j\in\mathcal{M}_{F}^{1}}\frac{(\mu_{1j}-\gamma_{j})^{2}}{2\sigma_{1j}^{2}}\\
		\leq &\min_{i\neq 1}\Bigg(\frac{(\mu_{i0}-\mu_{10})^{2}}{2(\sigma_{i0}^{2}/\alpha_{i}+\sigma_{10}^{2}/\tilde{\alpha}_{1})}\mathbf{1}\{i\in \mathcal{F}_{w}\cup \mathcal{I}_{w}\}
		+\alpha_{i}\sum\limits_{j\in\mathcal{M}_{I}^{i}}\frac{(\mu_{ij}-\gamma_{j})^{2}}{2\sigma_{ij}^{2}}\mathbf{1}\{i\in \mathcal{I}_{b}\cup \mathcal{I}_{w}\}\Bigg).
	\end{split}
\end{equation*}
If $\exists i\neq 1$ such that $\mathbb{P}\{FE_{i}\}\leq  \mathbb{P}\{FE_{1}\}$,
\begin{equation*}
	\begin{split}
		&\max_{\alpha_{1}\in(\tilde{\alpha}_{1},1)}\lim\limits_{n\rightarrow\infty}-\frac{1}{n}\log(1-P_{n,1})\\
		=&\max_{\alpha_{1}\in(\tilde{\alpha}_{1},1)}\Bigg(\min_{i\neq 1}\Big(\frac{(\mu_{i0}-\mu_{10})^{2}}{2(\sigma_{i0}^{2}/\alpha_{i}+\sigma_{10}^{2}/\alpha_{1})}\mathbf{1}\{i\in \mathcal{F}_{w}\cup \mathcal{I}_{w}\}
		+\alpha_{i}\sum\limits_{j\in\mathcal{M}_{I}^{i}}\frac{(\mu_{ij}-\gamma_{j})^{2}}{2\sigma_{ij}^{2}}\mathbf{1}\{i\in \mathcal{I}_{b}\cup \mathcal{I}_{w}\}\Big)\Bigg)\\
		\geq &\min_{i\neq 1}\bigg(\frac{(\mu_{i0}-\mu_{10})^{2}}{2(\sigma_{i0}^{2}/\alpha_{i}+\sigma_{10}^{2}/\tilde{\alpha}_{1})}\mathbf{1}\{i\in \mathcal{F}_{w}\cup \mathcal{I}_{w}\}
		+\alpha_{i}\sum\limits_{j\in\mathcal{M}_{I}^{i}}\frac{(\mu_{2j}-\gamma_{j})^{2}}{2\sigma_{ij}^{2}}\mathbf{1}\{i\in \mathcal{I}_{b}\cup \mathcal{I}_{w}\}\Bigg).
	\end{split}
\end{equation*}
Hence, 
\begin{equation*}
	\max_{\alpha_{1}\in(\tilde{\alpha}_{1},1)}\lim\limits_{n\rightarrow\infty}-\frac{1}{n}\log(1-P_{n,1})\geq \max_{\alpha_{1}\in(0,\tilde{\alpha}_{1}]}\lim\limits_{n\rightarrow\infty}-\frac{1}{n}\log(1-P_{n,1}).
\end{equation*}
Then when the rate of posterior convergence for the probability of false selection is optimal, $1-P_{n,1}\dot{=}\max_{i\in\{2,...,k\}}\mathbb{P}\{FE_{i}\}$, which indicates that, for this rate to be optimal, we just need to consider the case of $\mathbb{P}\{FE_1\}\leq \mathbb{P}\{FE_i\}$ for some $i\neq 1$. In this case,
\begin{equation}
	\begin{split}
		&1-P_{n,1}\\
		\dot{=}&\exp\Bigg(-n\min\limits_{i\in A}\bigg(\frac{(\mu_{n,i0}-\mu_{n,10})^{2}}{2(\sigma_{i0}^{2}n/N_{n,i}+\sigma_{10}^{2}n/N_{n,1})}\mathbf{1}\{i\in \mathcal{F}^{n}\cup \mathcal{I}_{w}^{n}\}
		+\sum\limits_{j\in\mathcal{M}_{I}^{i,n}}\frac{(\gamma_{j}-\mu_{n,ij})^{2}}{2\sigma_{ij}^{2}n/N_{n,i}}
		\mathbf{1}\{i\in \mathcal{I}_{b}^{n}\cup \mathcal{I}_{w}^{n}\}\bigg)\Bigg).
	\end{split}
\end{equation}

We have
\begin{equation*}
	1-P_{n,1}\dot{=}\max_{i\in\{1,2,\ldots,k\}}\mathbb{P}\{FE_{i}\}.
\end{equation*}
$\mathbb{P}\{FE_{1}\}$ and $\mathbb{P}\{FE_{i}\}$, $i\neq 1$ have been studied in Lemma 1.
Note that $V$ contains arms that are only pulled for finite times. First, suppose that $V$ is nonempty. For each $i\in A$, we define
\begin{equation*}
	\mu_{\infty,ij}\triangleq\lim\limits_{n\rightarrow\infty}\mu_{n,ij},\quad\quad\sigma_{\infty,ij}\triangleq\lim\limits_{n\rightarrow\infty}\sigma_{n,ij}.
\end{equation*} 
Recall that for each $i\in A$, the prior is $\mu_{1,ij}=0$ and $\sigma_{1,ij}=\infty$. Then, if $N_{n,i}=\sum\limits_{t=1}^{n-1}\mathbf{1}\{I_{t}=i\}=0$, $\mu_{n,ij}=\mu_{1,ij}=0$, $\sigma_{n,ij}=\sigma_{1,ij}=\infty$. If $N_{n,i}>0$,
\begin{equation*}
	\mu_{n,ij}=\frac{1}{N_{n,i}}\sum\limits_{t=1}^{n-1}\mathbf{1}\{I_{t}=i\}X_{t,ij},\quad\quad\sigma_{n,ij}^{2}=\frac{\sigma_{ij}^{2}}{N_{n,i}}.
\end{equation*}
Hence, for $i\in I$, $\mu_{\infty,ij}=\mu_{ij}$ and $\sigma_{\infty,ij}^{2}=0$ while for $i \in V$, $\sigma_{\infty,ij}>0$.

For arm $i$ in the nonempty set $V$, we have $P_{\infty,i}\in(0,1)$ because $\sigma_{\infty,i}>0$ and $\sigma_{\infty,ij}>0$. This implies $P_{\infty,1}<1$ and so $\lim\limits_{n\rightarrow\infty}-\frac{1}{n}\log(1-P_{n,1})=\lim\limits_{n\rightarrow\infty}-\frac{1}{n}\log(1-P_{\infty,1})=0$.

Now suppose that $V$ is empty. 
\begin{equation}
	\begin{split}
		&1-P_{n,1}\\
		=&\mathbb{P}_{\theta\sim \Pi_{n}}\bigg(FE_{1}\cup\bigcup\limits_{i\neq1}FE_{i}\bigg)\\
		\dot{=}&\exp\Bigg(-n\min\Bigg(\min\limits_{i\neq1}\Big(\frac{(\mu_{n,i0}-\mu_{n,10})^{2}}{2(\sigma_{i0}^{2}n/N_{n,i}+\sigma_{10}^{2}n/N_{n,1})} \mathbf{1}\{i\in\mathcal{F}_{w}^{n}\cup \mathcal{I}_{w}^{n}\}
		+\sum\limits_{j\in\mathcal{M}_{I}^{i,n}}\frac{(\gamma_{j}-\mu_{n,ij})^{2}}{2\sigma_{ij}^{2}n/N_{n,i}}\mathbf{1}\{i\in\mathcal{I}_{b}^{n}\cup \mathcal{I}_{w}^{n}\}\Big),\\
		&\min\limits_{j\in\mathcal{M}_{F}^{1,n}}\frac{(\gamma_{j}-\mu_{n,1j})^{2}}{2\sigma_{1j}^{2}n/N_{n,1}}\bigg)\Bigg).
	\end{split}
\end{equation}
Under any sampling rule,
\begin{equation}
	\begin{split}
		&1-P_{n,1}\\
		\geq&\exp\Bigg(-n\max_{\alpha\in \Omega}\min\bigg(\min\limits_{i\neq1}\Big(\frac{(\mu_{n,i0}-\mu_{n,10})^{2}}{2(\sigma_{i0}^{2}/\alpha_{i}+\sigma_{10}^{2}/\alpha_{1})}\mathbf{1}\{i\in\mathcal{F}_{w}^{n}\cup \mathcal{I}_{w}^{n}\}
		+\sum\limits_{j\in\mathcal{M}_{I}^{i,n}}\frac{(\gamma_{j}-\mu_{n,ij})^{2}}{2\sigma_{ij}^{2}/\alpha_{i}}\mathbf{1}\{i\in\mathcal{I}_{b}^{n}\cup \mathcal{I}_{w}^{n}\} \Big),\\ &\min\limits_{j\in\mathcal{M}_{F}^{1,n}}\frac{(\gamma_{j}-\mu_{n,1j})^{2}}{2\sigma_{1j}^{2}/\alpha_{1}}\bigg)\Bigg).
	\end{split}
\end{equation}
Since every arm will be pulled infinitely as $n\rightarrow\infty$, $\mu_{n,i0}\rightarrow\mu_{i0}$ and $\mu_{n,ij}\rightarrow \mu_{ij}$ for $i\in\{1,2,\ldots,k\}$ and $j\in\{1,2,\ldots,m\}$. Thus proving
\begin{equation}
	\lim\limits_{n\rightarrow\infty}\sup-\frac{1}{n}\log(1-P_{n,1})\leq\Gamma_{\beta^{*}}.
\end{equation}
and for $\beta\in(0,1)$, under any sampling rule satisfying $N_{n,1}/n\rightarrow\beta$,
\begin{equation*}
	\lim\sup\limits_{n\rightarrow\infty}-\frac{1}{n}\log (1-P_{n,1})\leq\Gamma_{\beta},
\end{equation*}
are equivalent to show
\begin{equation}\label{star}
	\begin{split}
		\Gamma_{\beta^{*}}=&\max_{\alpha\in \Omega}\min\Bigg(\min\limits_{i\neq1}\bigg(\frac{(\mu_{i0}-\mu_{10})^{2}}{2(\sigma_{i0}^{2}/\alpha_{i}+\sigma_{10}^{2}/\alpha_{1})} \mathbf{1}\{i\in\mathcal{F}_{w}\cup \mathcal{I}_{w}\}+\sum\limits_{j\in\mathcal{M}_{I}^{i}}\frac{(\gamma_{j}-\mu_{ij})^{2}}{2\sigma_{ij}^{2}/\alpha_{i}}\mathbf{1}\{i\in\mathcal{I}_{b}\cup \mathcal{I}_{w}\}\bigg),\\
        &\min\limits_{j\in\mathcal{M}_{F}^{1}}\frac{(\gamma_{j}-\mu_{1j})^{2}}{2\sigma_{1j}^{2}/\alpha_{1}}\Bigg)
	\end{split}
\end{equation}
and 
\begin{equation}\label{beta}
	\begin{split}
		\Gamma_{\beta}=&\max_{\alpha\in \Omega: \alpha_{1}=\beta}\min\Bigg(\min\limits_{i\neq1}\bigg(\frac{(\mu_{i0}-\mu_{10})^{2}}{2(\sigma_{i0}^{2}/\alpha_{i}+\sigma_{10}^{2}/\alpha_{1})}\mathbf{1}\{i\in\mathcal{F}_{w}\cup \mathcal{I}_{w}\}+\sum\limits_{j\in\mathcal{M}_{I}^{i}}\frac{(\gamma_{j}-\mu_{ij})^{2}}{2\sigma_{ij}^{2}/\alpha_{i}}\mathbf{1}\{i\in\mathcal{I}_{b}\cup \mathcal{I}_{w}\}\bigg),\\
        &\min\limits_{j\in\mathcal{M}_{F}^{1}}\frac{(\gamma_{j}-\mu_{1j})^{2}}{2\sigma_{1j}^{2}/\alpha_{1}}\Bigg),
	\end{split}
\end{equation}
where $\beta\in (0,1)$. 

Let $r_{i}(\beta,\alpha_{i})=\frac{(\mu_{i0}-\mu_{10})^{2}}{2(\sigma_{i0}^{2}/\alpha_{i}+\sigma_{10}^{2}/\alpha_{1})}\mathbf{1}\{i\in\mathcal{F}_{w}\cup \mathcal{I}_{w}\}+\sum\limits_{j\in\mathcal{M}_{I}^{i}}\frac{(\gamma_{j}-\mu_{ij})^{2}}{2\sigma_{ij}^{2}/\alpha_{i}}\mathbf{1}\{i\in\mathcal{I}_{b}\cup \mathcal{I}_{w}\}$, where $i\neq 1$. Recall that
\begin{equation*}
	\begin{split}
		&\frac{(\mu_{i0}-\mu_{10})^{2}}{(\sigma_{i0}^{2}/\alpha_{i}^{\beta}+\sigma_{10}^{2}/\beta)}\mathbf{1}\{i\in\mathcal{F}_{w}\cup \mathcal{I}_{w}\}
		+\alpha_{i}^{\beta}\sum\limits_{j\in\mathcal{M}_{I}^{i}}\frac{(\mu_{ij}-\gamma_{j})^{2}}{\sigma_{ij}^{2}}\mathbf{1}\{i\in\mathcal{I}_{b}\cup \mathcal{I}_{w}\}\\
		=&\frac{(\mu_{i'0}-\mu_{10})^{2}}{(\sigma_{i'0}^{2}/\alpha_{i'}^{\beta}+\sigma_{10}^{2}/\beta)}\mathbf{1}\{i'\in\mathcal{F}_{w}\cup \mathcal{I}_{w}\}
		+\alpha_{i'}^{\beta}\sum\limits_{j\in \mathcal{M}_{I}^{i'}}\frac{(\mu_{i'j}-\gamma_{j})^{2}}{\sigma_{i'j}^{2}}\mathbf{1}\{i'\in\mathcal{I}_{b}\cup \mathcal{I}_{w}\},
	\end{split}
\end{equation*}
i.e. $r_{i}(\beta,\alpha_{i}^{\beta})=r_{i'}(\beta,\alpha_{i'}^{\beta})$, where $i,i'=2,3,\ldots,k$ and $i\neq i'$. We prove (\ref{star}) and (\ref{beta}) by contradiction. Suppose there exists $\alpha'\neq \alpha_{i}^{\beta}\in\Omega:\alpha_{1}=\beta$ satisfying $\max_{\alpha\in\Omega:\alpha_{1}=\beta}\min_{i\neq 1}r_{i}(\beta,\alpha_{i})=\min_{i\neq 1}r_{i}(\beta,\alpha_{i}')$. Since the solution to $r_{i}(\beta,\alpha_{i})=r_{i'}(\beta,\alpha_{i'})$ and $\sum_{i=2}^{k}\alpha_{i}=1-\beta$ is unique, there must exist some $i'$ such that $ r_{i'}(\beta,\alpha_{i'}')> \min_{i\neq 1}r_{i}(\beta,\alpha_{i}')$. Since $r_{i}(\beta, \alpha_{i})$ is continuous, we consider a new vector $\alpha^{\epsilon}$ with $\alpha_{i'}^{\epsilon}=\alpha_{i'}'-\epsilon$ and $\alpha_{i}^{\epsilon}=\alpha_{i}'+\epsilon/(k-2)$ for $i\neq i'$ and $i\neq 1$. For sufficiently small $\epsilon$, we have
\begin{equation*}
	r_{i'}(\beta, \alpha_{i'}^{\epsilon})>\min_{i\neq 1}r_{i}(\beta,\alpha_{i}^{\epsilon})>\min_{i\neq 1}r_{i}(\beta,\alpha_{i}').
\end{equation*}
Then $\min_{i\neq 1}r_{i}(\beta, \alpha_{i}^{\epsilon})>\max_{\alpha\in\Omega:\alpha_{1}=\beta}\min_{i\neq 1}r_{i}(\beta,\alpha_{i})$. This yields a contradiction. Since $\alpha$ will not affect $\min\limits_{j\in\mathcal{M}_{F}^{1}}\frac{(\gamma_{j}-\mu_{1j})^{2}}{2\sigma_{1j}^{2}/\beta}$ when $\beta$ is given, $\Gamma_{\beta}=\min(\min_{i\neq 1}r_{i}(\beta,\alpha_{i}^{\beta}), \min\limits_{j\in\mathcal{M}_{F}^{1}}\frac{(\gamma_{j}-\mu_{1j})^{2}}{2\sigma_{1j}^{2}/\beta})$.
Similarly, we can get the result when $\beta=\beta^{*}$. 

If $N_{n,i}/n\rightarrow \alpha_{i}^{*}$ for each $i\in A$, by (5), for each $i\neq1$, we have
\begin{equation}
	\begin{split}
		\lim\limits_{n\rightarrow\infty}\frac{(\mu_{n,i0}-\mu_{n,10})^{2}}{2(\sigma_{i0}^{2}n/N_{n,i}+\sigma_{10}^{2}n/N_{n,1})} \mathbf{1}\{i\in\mathcal{F}_{w}^{n}\cup \mathcal{I}_{w}^{n}\} +\sum\limits_{j\in\mathcal{M}_{I}^{i,n}}\frac{(\gamma_{j}-\mu_{n,ij})^{2}}{2\sigma_{ij}^{2}n/N_{n,i}}\mathbf{1}\{i\in\mathcal{I}_{b}^{n}\cup \mathcal{I}_{w}^{n}\} \\
		=\frac{(\mu_{i0}-\mu_{10})^{2}}{2(\sigma_{i0}^{2}/\alpha_{i}^{*}+\sigma_{10}^{2}/\beta^{*})} \mathbf{1}\{i\in\mathcal{F}_{w}\cup \mathcal{I}_{w}\}+\sum\limits_{j\in\mathcal{M}_{I}^{i}}\frac{(\gamma_{j}-\mu_{ij})^{2}}{2\sigma_{ij}^{2}/\alpha_{i}^{*}}\mathbf{1}\{i\in\mathcal{I}_{b}\cup \mathcal{I}_{w}\}=\Gamma_{\beta^{*}}
	\end{split}
\end{equation}
and thus 
\begin{equation}
	\begin{split}
		&1-P_{n,1}\\
		\dot{=}&\exp\Bigg(-n\min\bigg(\min\limits_{i\neq1}\Big(\frac{(\mu_{n,i0}-\mu_{n,10})^{2}}{2(\sigma_{i0}^{2}n/N_{n,i}+\sigma_{10}^{2}n/N_{n,1})} \mathbf{1}\{i\in\mathcal{F}_{w}^{n}\cup \mathcal{I}_{w}^{n}\}
		+\sum\limits_{j\in\mathcal{M}_{I}^{i,n}}\frac{(\gamma_{j}-\mu_{n,ij})^{2}}{2\sigma_{ij}^{2}n/N_{n,i}}\mathbf{1}\{i\in\mathcal{I}_{b}^{n}\cup \mathcal{I}_{w}^{n}\}\Big),\\ &\min\limits_{j\in\mathcal{M}_{F}^{1,n}}\frac{(\gamma_{j}-\mu_{n,1j})^{2}}{2\sigma_{1j}^{2}n/N_{n,1}}\bigg)\Bigg)\\
		\dot{=}&\exp\big(-n\Gamma_{\beta^{*}}\big),
	\end{split}
\end{equation}
which implies
\begin{equation*}
	\lim\limits_{n\rightarrow\infty}-\frac{1}{n}\log(1-P_{n,1})=\Gamma_{\beta^{*}}.
\end{equation*}
Similarly, for $\beta\in(0,1)$, under any sampling rule satisfying $\frac{N_{n,i}}{n}\rightarrow \alpha_{i}^{\beta}$, $i \in A$,
\begin{equation*}
	\lim\limits_{n\rightarrow\infty}-\frac{1}{n}\log (1-P_{n,1})=\Gamma_{\beta}.
\end{equation*}
\end{document}